\documentclass[a4paper,fleqn]{cas-dc}
\usepackage{orcidlink}
\usepackage{threeparttable}
\usepackage{booktabs}
\usepackage[numbers]{natbib}
\def\tsc#1{\csdef{#1}{\textsc{\lowercase{#1}}\xspace}}
\tsc{WGM}
\tsc{QE}
\tsc{EP}
\tsc{PMS}
\tsc{BEC}
\tsc{DE}
\begin{document}
\let\WriteBookmarks\relax
\def\floatpagepagefraction{1}
\def\textpagefraction{.001}
\let\printorcid\relax
\shorttitle{Ronghui Zhang et~al.}
\shortauthors{Ronghui Zhang et~al.}

\title [mode = title]{ABCDWaveNet: Advancing Robust Road Ponding Detection in Fog through Dynamic Frequency-Spatial Synergy}
% A Dynamic Multi-Scale Deep Learning Framework for Robust Road Ponding Detection under Adverse Foggy Conditions}                      
%\tnotemark[1]

\tnotetext[1]{This project is jointly supported by the National Natural Science Foundation
of China (Nos.52172350, W2421069 and 51775565), the Guangdong Basic and Applied Research Foundation (No. 2022B1515120072), the
Guangzhou Science and Technology Plan Project (No.2024B01W0079), the Nansha Key RD Program (No.2022ZD014), the Science and Technology Planning Project of Guangdong Province (No.2023B1212060029), and the China Postdoctoral Science Foundation(No.2013T60904)}

%\tnotetext[2]{The second title footnote which is a longer text matter
   %to fill through the whole text width and overflow into
   %another line in the footnotes area of the first page.}

\author[1]{Ronghui Zhang\orcidlink{0000-0001-6107-4044}}
\ead{zhangrh25@mail.sysu.edu.cn}
\author[1]{Dakang Lyu\orcidlink{0009-0004-9396-7735}}
\ead{lvdk@mail2.sysu.edu.cn}

\author[1]{Tengfei Li\orcidlink{0009-0009-9579-0044}}
%\fnmark[2]
\ead{litf23@mail2.sysu.edu.cn}
%\ead[URL]{https://www.university.org}

\author[1]{Yunfan Wu\orcidlink{0009-0004-8938-1694}}
\ead{wuyf227@mail2.sysu.edu.cn}

%\credit{Data curation, Writing - Original draft preparation}
\author[2,3]{Ujjal MANANDHAR\orcidlink{0000-0003-2669-2095}}
\ead{ujjal001@e.ntu.edu.sg}

\author[1]{Benfei Wang\orcidlink{0000-0001-9270-4002}}
\ead{wangbf8@mail.sysu.edu.cn}

\author[1]{Junzhou Chen\orcidlink{0000-0002-3388-3503}}
\cormark[1]
\ead{chenjunzhou@mail.sysu.edu.cn}

\author[4]{Bolin Gao\orcidlink{0000-0002-5582-7289}}
\ead{gaobolin@tsing.edu.cn}

\author[2]{Danwei Wang\orcidlink{0000-0003-3400-0079}}
\ead{edwwang@ntu.edu.sg}

\author[5]{Yiqiu Tan\orcidlink{0000-0002-8557-9116}}
\ead{tanyiqiu@hit.edu.cn}

\cortext[cor1]{Corresponding author}
%\cortext[cor2]{Principal corresponding author}

%\credit{Conceptualization of this study, Methodology, Software}

%\address[1]{, Street 129, 1043 NX Amsterdam, The Netherlands}
\affiliation[1]{organization={Guangdong Provincial Key Laboratory of Intelligent Transportation System, School of Intelligent Systems Engineering},
                addressline={Sun Yat-sen University}, 
                city={Guangzhou},
%               citysep={}, % Uncomment if no comma needed between city and postcode
                postcode={510275}, 
                %state={Kerala},
                country={China}}
\affiliation[2]{organization={School of Electrical and Electronic Engineering},
                addressline={Nanyang Technological University}, 
                %city={Guangzhou},
%               citysep={}, % Uncomment if no comma needed between city and postcode
                postcode={639798}, 
                %state={Kerala},
                country={Singapore}}
\affiliation[3]{organization={Western Power},
                %addressline={The University of Western Australia}, 
                city={Perth WA},
%               citysep={}, % Uncomment if no comma needed between city and postcode
                postcode={6000}, 
                %state={Kerala},
                country={Australia}}
\affiliation[4]{organization={School of Vehicle and Mobility},
                addressline={Tsinghua University}, 
                city={Beijing},
%               citysep={}, % Uncomment if no comma needed between city and postcode
                postcode={150090}, 
                %state={Kerala},
                country={China}}
\affiliation[5]{organization={School of Transportation Science and Engineering},
                addressline={Harbin Institute of Technology}, 
                city={Harbin},
%               citysep={}, % Uncomment if no comma needed between city and postcode
                postcode={10084}, 
                %state={Kerala},
                country={China}}

\begin{abstract}
Road ponding presents a significant threat to vehicle safety, particularly in adverse fog conditions, where reliable detection remains a persistent challenge for Advanced Driver Assistance Systems (ADAS). To address this, we propose ABCDWaveNet, a novel deep learning framework leveraging Dynamic Frequency-Spatial Synergy for robust ponding detection in fog. The core of ABCDWaveNet achieves this synergy by integrating dynamic convolution for adaptive feature extraction across varying visibilities with a wavelet-based module for synergistic frequency-spatial feature enhancement, significantly improving robustness against fog interference. Building on this foundation, ABCDWaveNet captures multi-scale structural and contextual information, subsequently employing an Adaptive Attention Coupling Gate (AACG) to adaptively fuse global and local features for enhanced accuracy. To facilitate realistic evaluations under combined adverse conditions, we introduce the Foggy Low-Light Puddle dataset. Extensive experiments demonstrate that ABCDWaveNet establishes new state-of-the-art performance, achieving significant Intersection over Union (IoU) gains of 3.51\%, 1.75\%, and 1.03\% on the Foggy-Puddle, Puddle-1000, and our Foggy Low-Light Puddle datasets, respectively. Furthermore, its processing speed of 25.48 FPS on an NVIDIA Jetson AGX Orin confirms its suitability for ADAS deployment. These findings underscore the effectiveness of the proposed Dynamic Frequency-Spatial Synergy within ABCDWaveNet, offering valuable insights for developing proactive road safety solutions capable of operating reliably in challenging weather conditions.
\end{abstract}

%%%%%%%%%\begin{graphicalabstract}
%%%%%%%%%\includegraphics{figs/cas-grabs.pdf}
%%%\end{graphicalabstract}

%\begin{highlights}
%\item Proposes ABCDWaveNet for robust ponding detection under foggy conditions.

%\item Integrates dynamic convolution and wavelet transform for adaptive feature fusion.

%\item Designs adaptive multi-scale fusion (ABC mechanism) for accurate detection.

%\item Achieves SOTA performance on multiple benchmarks, enabling early warning of road ponding in foggy scenarios.
%\end{highlights}

\maketitle

\section{Introduction}

In contemporary transportation, the rapid increase of traffic volume driven by urbanization has heightened the demand for road safety under all-weather conditions. Despite notable advances in vehicle safety and traffic infrastructure, maintaining road safety during adverse weather, especially in foggy conditions, remains a persistent and critical challenge. Fog, one of the most common weather phenomena, drastically reduces visibility, a critical element of safe driving \cite{ali2024effect, engineeringSurvey}. Under foggy conditions, drivers struggle to accurately perceive their surroundings, which encompass not only other vehicles and pedestrians but also potential road hazards, such as road ponding and potholes. 

This reduced visibility markedly increases the risks associated with these hazards. For example, under normal conditions, standing water on the road can cause vehicles to skid and extend braking distances. In fog, these risks are intensified, as drivers find it increasingly difficult to identify road ponding in a timely manner, potentially resulting in severe consequences for both life and property \cite{yan2023intelligent, engineeringAI}. The US Federal Highway Administration (FHWA) reports that over 38,700 vehicle accidents occur annually in foggy conditions, leading to more than 600 fatalities and over 16,300 injuries \cite{FHWA2023}. Research indicates that the probability of accidents in fog is 35 times greater than in clear weather, and adverse weather conditions account for approximately 20\% of all traffic incidents. Furthermore, these conditions contribute to 38.3\% of congestion and 23\% of non-recurring delays, resulting in billions of dollars in economic losses \cite{Gong2022}. 

\begin{figure*}[htbp]  % h: here, t: top, b: bottom, p: page of floats
    \centering
    \includegraphics[width = 0.9\textwidth]{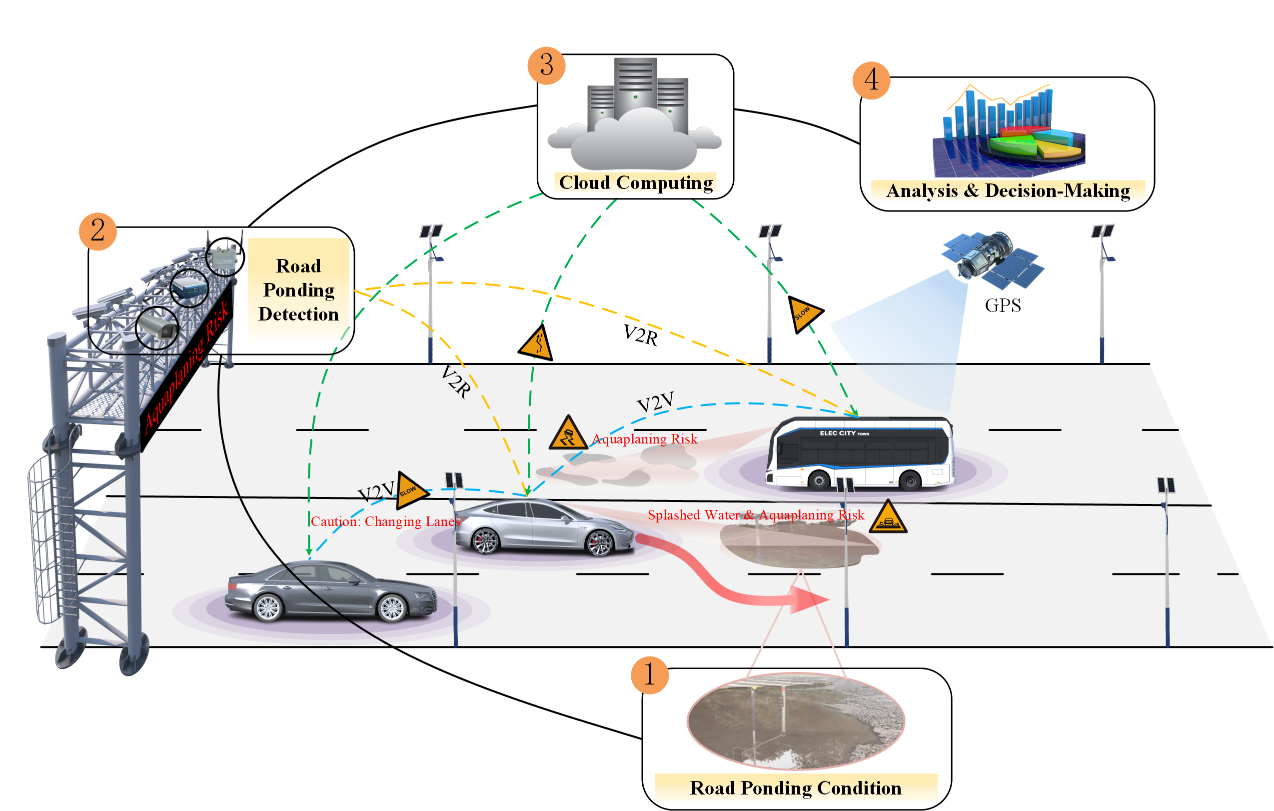}  
    \caption{Road Ponding Detection in an Advanced Driver Assistance System (ADAS) \cite{zhaonan}.}
    \label{fig:adas} 
\end{figure*}

To address these challenges, advanced driver assistance system (ADAS) have emerged, which integrate various traffic-related data to facilitate safer, more efficient and environmentally sustainable transportation. Fig. \ref{fig:adas} illustrates the road ponding detection process in an ADAS system under foggy conditions \cite{Engineering+ICV, Engineering+road}. However, traditional sensor-based methods encounter significant limitations in fog-prone environments \cite{sensors}. Sensors, such as hydrogel sensors \cite{hydrogel} and infrared devices \cite{infrared}, are susceptible to signal distortion caused by fog, which compromises their detection accuracy. Furthermore, their limited spatial coverage hampers comprehensive monitoring of road conditions.
In contrast, deep learning has emerged as a powerful tool in various domains, particularly in visual analysis and feature extraction \cite{engineeringCV}. The capacity of deep learning models, such as convolutional neural networks (CNNs) and transformers \cite{attention, ABNet}, to learn hierarchical features renders them especially effective for complex visual tasks. This capability is particularly relevant for monitoring road conditions, as it enables the detection of subtle visual cues indicative of adverse weather effects, which traditional methods often struggle to identify. Furthermore, the application of deep learning in this context has shown promise in enhancing road safety and traffic management. However, challenges such as ensuring model robustness in diverse environmental conditions remain critical considerations \cite{li2024efficient, engineering+transportation}.

By harnessing deep learning techniques, we can significantly improve the detection of road hazards in foggy conditions, thereby enhancing overall road safety. In this paper, we propose the \underline{A}ggregation-\underline{B}roadcast-\underline{C}oupling \underline{D}ynamic \underline{Wave}let \underline{Net}work (ABCDWaveNet), a novel deep learning framework specifically designed to overcome the challenges of detecting road ponding in foggy conditions. ABCDWaveNet combines dynamic convolution , wavelet-based feature extraction, and a multi-scale feature fusion strategy to enhance robustness and accuracy in adverse environments. The architecture is designed to adaptively capture critical spatial features while mitigating the impact of fog-induced interference. Through its Aggregation-Broadcast-Coupling mechanism, ABCDWaveNet efficiently fuses global and local information, improving feature representation across varying scales and leading to more precise detection of road ponding, providing reference and inspiration for advancing the field of road safety under adverse weather conditions. The key contributions of this paper are as follows:

1) \textbf{Novel Framework Design:} We introduce ABCDWaveNet, an innovative deep learning architecture that integrates dynamic convolution with advanced feature fusion mechanisms to enhance road ponding detection under foggy conditions. Unlike traditional convolutional architectures with fixed kernels, ABCDWaveNet employs dynamic convolution, enhancing its adaptability to non-uniform visual patterns caused by varying environmental conditions. By capturing both spatial and frequency domain features, the framework significantly improves robustness and accuracy in challenging foggy environments.

2) \textbf{Incorporation of Discrete Wavelet Transform:} ABCDWaveNet incorporates the Discrete Wavelet Transform (DWT) in its encoding stages, effectively decomposing features into low- and high-frequency sub-bands. This decomposition enables the model to capture global structural patterns while preserving intricate details, thereby mitigating fog-induced interference and enhancing feature robustness.

3) \textbf{Multi-Scale Feature Integration:} Our Aggregation-Broadcast-Coupling Mechanism combines multi-scale information aggregation with an Adaptive Attention Coupling Gate (AACG), facilitating efficient integration of global and local features. This mechanism dynamically regulates information flow, allowing the model to adapt to diverse ponding scales while maintaining high decoding accuracy.

4) \textbf{Dataset Development and Performance:} To address the lack of datasets simulating realistic adverse weather conditions, we developed the Foggy Low-Light Puddle benchmark dataset, featuring low light, dense fog, and diverse ponding forms. Extensive experiments demonstrate that ABCDWaveNet achieves state-of-the-art performance, with Intersection over Union (IoU) improvements of 3.51\%, 1.75\%, and 1.03\% on the Foggy-Puddle, Puddle-1000, and Foggy Low-Light Puddle datasets, respectively. Additionally, ABCDWaveNet achieves 25.48 FPS on an NVIDIA Jetson AGX Orin, making it suitable for deployment in ADAS for early road ponding warnings under foggy conditions.

\section{Related Work}
\subsection{Sensor-based methods}
Sensor-based methods, such as LiDAR \cite{lidar}, hydrogel sensors \cite{hydrogel}, and infrared devices \cite{infrared,near-infrared}, are widely used for road surface water detection, offering high accuracy under various conditions. For instance, LiDAR-based algorithms leverage histogram analysis of point cloud data for real-time segmentation and water hazard detection \cite{lidar}. Similarly, near-infrared optoelectronic techniques exploit light reflection, scattering, and polarization properties to classify road surfaces \cite{near-infrared}.

However, these methods face critical limitations, particularly in foggy conditions. Fog-induced signal distortion significantly reduces the effectiveness of hydrogel sensors \cite{hydrogel} and infrared devices \cite{infrared}, while LiDAR performance is hindered by diminished visibility \cite{lidar}. Furthermore, the limited spatial coverage of these sensors, combined with the high costs and operational complexity of specialized hardware, restrict their scalability and practicality for large-scale or adverse-weather applications.

\subsection{Machine learning-based Methods}
\subsubsection{Classical Machine Learning Methods}
Classical machine learning methods, including Support Vector Machines (SVMs) and Random Forests, have been utilized for water detection by extracting hand-crafted features such as color and texture. Optical flow techniques combined with SVMs \cite{svm1} and SVM-based hypothesis generation approaches using color information \cite{svm2} have been proposed to detect wet road surfaces. Optimized SVMs with image segmentation \cite{svm3,roadsurface} and Random Forest classifiers based on HSV color models \cite{RandomForest} have also shown effectiveness in controlled settings. For feature enhancement, wavelet packet transforms were integrated with SVMs to improve wet surface classification \cite{svm4}. While these approaches are effective in simple environments, they are sensitive to noise and struggle with dynamically changing conditions, particularly in foggy scenarios where visual distortions compromise hand-crafted feature reliability \cite{Engineering+Rev}.

\subsubsection{Deep Learning-based Methods}
Deep learning methods, particularly convolutional neural networks (CNNs), have advanced water detection, achieving state-of-the-art performance\cite{engineeringRain}. U-Net \cite{unet}, with its U-shaped encoder-decoder structure, has become foundational in segmentation tasks, outperforming traditional HSV-based methods \cite{ICARES}. Extensions like U-Net-RAU \cite{waterhazard} enhance pixel-level segmentation by leveraging water reflection properties, while SWNet \cite{swnet} uses multiscale feature fusion and splash attention modules for improved water splash detection. AGSENet \cite{agsenet} further incorporates multi-scale attention mechanisms for enhanced feature discrimination, addressing more complex water detection scenarios.

Transformer-based models, such as Vision Transformers (ViT) 
\cite{vit, engineeringVit}, have gained traction in semantic segmentation. SeaFormer \cite{seaformer}, a lightweight Transformer architecture, demonstrates strong performance in efficiently capturing global context and local spatial details. However, challenges remain in computational efficiency and edge detail accuracy, which are critical for precise water segmentation tasks.
\subsection{Foggy Conditions Detection}
Foggy conditions exacerbate ponding detection challenges due to reduced visibility and contrast \cite{adversaial}. Traditional dehazing methods, such as the Dark Channel Prior (DCP) \cite{dcp}, enhance visual clarity but often fail to retain fine details, limiting their applicability to complex segmentation tasks. Deep learning models, such as AOD-Net \cite{AOD-Net}, offer real-time dehazing but are not specifically designed for water detection. Recent segmentation approaches address fog-specific challenges with domain-adaptive methods. FIFO \cite{FIFO} employs fog-pass filtering to improve feature consistency across foggy domains, while CuDA-Net \cite{cudanet} disentangles fog-related and style-related influences to enhance domain adaptation. Bi-directional Wavelet Guidance (BWG) \cite{bwg} refines feature extraction by decoupling style features through wavelet transformations, preserving crucial details and improving segmentation robustness. Its bi-directional structure integrates local and global contextual information, making it effective in foggy environments.

While these advancements address fog-related challenges, further research is needed to integrate dehazing techniques with task-specific segmentation frameworks for robust road ponding detection under adverse weather conditions.

\section{Proposed Method}
In this section, we begin by presenting an overview of the
network’s architecture. Subsequently, we provide an in-depth explanation of the individual modules within the network.
\subsection{Overall Architecture}
\begin{figure*}[htbp]  % h: here, t: top, b: bottom, p: page of floats
    \centering
    \includegraphics[width=1\textwidth]{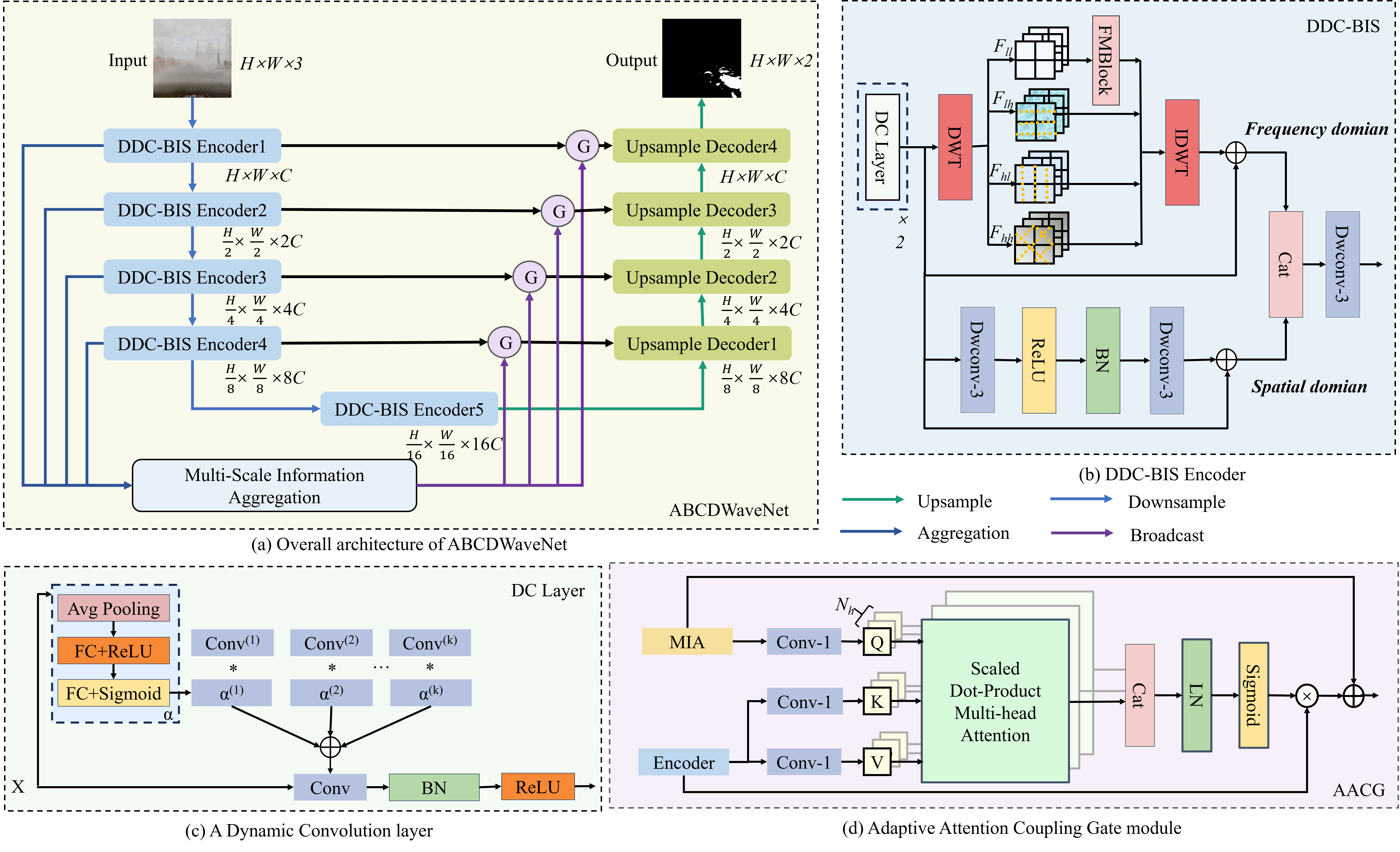}  
    \caption{The Architecture of AGSENet(a): A U-Shaped Network Featuring Five Encoders, Four Decoders, Multi-Scale Information Aggregation (MIA), and Adaptive Attention Coupling Gate (AACG) Modules (d). The Encoder-Decoder Structure is Built by Stacking Dual Dynamic Convolution-Bidomain Information Synergy (DDC-BIS) Modules (b), with Each DDC Layer Comprising Two Consecutive Dynamic Convolution Layers (c).}
    \label{fig:over_architecture} 
\end{figure*}

Fig. \ref{fig:over_architecture} illustrates the comprehensive architecture of the proposed approach, ABCDWaveNet. Our ABCDWaveNet has a highly hierarchical
architecture (i.e., U-Net \cite{unet} like architecture), complemented
by skip connections.

Given the RGB image \( I \in \mathbb{R}^{H \times W \times C} \), where \( H \times W \) represents the spatial resolution and \( C \) is the number of channels, the ABCDWaveNet initiates feature extraction through a series of encoding stages that utilize a Dual Dynamic Convolutional layer and a Bidomain Information Synergy module. These components collaborate harmoniously to extract the output features. Next, The architecture integrates features from each encoding stages through the Multi-scale Information Aggregation module, facilitating comprehensive multi-scale representation. Running in a coarse-to-fine manner, the MIA module employs a multi-scale adaptive scale selection module designed to effectively capture essential structural and contextual information critical for detecting ponding under foggy conditions. This includes recognizing the contours of ponding edges and distinguishing the contrasting textures of the surrounding dry road surfaces. Subsequently, a progressive refinement module is implemented to distill the most discriminative features that significantly aid in differentiating ponding areas from non-ponding ones, focusing on variations in color and texture. The aggregated global information from the MIA module is then broadcast to each decoder through Adaptive Attention Coupling Gate (AACG) embedded in the skip connections. The AACG selectively fuses global and local information, allowing the model to highlight key structure and context-specific details. In the following subsections, we delve into the details of the newly introduced components.

\subsection{Dynamic Feature Extraction}
\subsubsection{Dynamic Convolution}
Extracting meaningful features for accurate ponding detection under adverse environmental conditions, such as fog, presents significant challenges. Foggy conditions severely degrade image contrast and obscure edge details, impairing the performance of traditional convolutional neural networks (CNNs) \cite{agsenet,ABNet,swnet}. These models often struggle to capture critical visual information effectively in such complex scenarios. To address these limitations, we introduce a Double Dynamic Convolution Layer in the encoding stage, designed to enhance the network's adaptability and robustness by employing dynamic feature extraction.

The Double Dynamic Convolution Layer utilizes dynamic convolution, which enhances representational capacity by dynamically combining multiple convolutional kernels based on the input's characteristics. As illustrated in Fig. \ref{fig:over_architecture} (c), dynamic convolution adaptively weights \( K \) parallel convolutional kernels \( \{W^{(k)}\}_{k=1}^{K} \) with input-dependent attention weights \( \alpha^{(k)} \). For an input feature map \( X \in \mathbb{R}^{H \times W \times C} \), the output is computed as:

\begin{equation}
\label{dyconv}
Y = \sum_{k=1}^{K} \alpha^{(k)} \cdot (X * W^{(k)}),
\end{equation}

where \( * \) denotes the convolution operation. The coefficient \( \alpha^{(k)} \) is dynamically generated based on the input using a MLP module. For a given input \( X \), global average pooling is applied to condense the spatial information into a vector. This vector is then processed by the MLP module with a Sigmoid activation to produce input-dependent coefficients:
\begin{equation}
\label{alpha}
\alpha = \text{Sigmoid}(\text{MLP}(\text{Pool}(X))),
\end{equation}

\textbf{Complexity Analysis.} We analyze the parameter count and FLOPs of dynamic convolution compared to standard convolution\cite{paramnet}. In standard convolution, each layer employs a single kernel, resulting in a parameter count of \( C_{\text{out}} \cdot C_{\text{in}} \cdot K \cdot K \), and the associated FLOPs are given by:

\begin{equation}
\label{flops_conv}
\text{FLOPs}_{\text{conv}} = H' \cdot W' \cdot C_{\text{out}} \cdot C_{\text{in}} \cdot K \cdot K,
\end{equation}

In contrast, dynamic convolution involves multiple kernels and an attention mechanism. The parameter count is calculated as \( C_{\text{in}}^2 + C_{\text{in}} \cdot M + M  \cdot C_{\text{out}} \cdot C_{\text{in}}  \cdot K^2 \), and the FLOPs for dynamic convolution are expressed as:

\begin{equation}
\label{flops_dy}
\text{FLOPs}_{\text{dy}} = C_{\text{in}}^2 + C_{\text{in}}  M + M C_{\text{out}}  C_{\text{in}} K^2 + H'  W'  C_{\text{out}}  C_{\text{in}} K^2,
\end{equation}
The parameter ratio of dynamic convolution over standard:
\begin{equation}
\begin{aligned}
R_{\text{param}} &= \frac{C_{\text{in}}^2 + C_{\text{in}} M + M C_{\text{out}} C_{\text{in}} K^2}{C_{\text{out}} C_{\text{in}} K K} \\
&= \frac{C_{\text{in}}}{C_{\text{out}} K^2} + \frac{M}{C_{\text{out}} K^2} + M \\
&\approx \frac{1}{K^2} + M, \quad (M \ll C_{\text{out}} K^2, \; C_{\text{in}} \approx C_{\text{out}})
\end{aligned}
\end{equation}
The FLOPs ratio is
\begin{equation}
\begin{aligned}
\label{r-flops}
R_{\text{FLOPs}} &= \frac{C_{\text{in}}^2 + C_{\text{in}} M + M C_{\text{out}} C_{\text{in}} K^2 + H' W' C_{\text{out}} C_{\text{in}} K^2}{H' W' C_{\text{out}} C_{\text{in}} K^2} \\
&= \frac{C_{\text{in}}}{H' W' C_{\text{out}} K^2} + \frac{M}{H' W' C_{\text{out}} K^2} + \frac{M}{H' W'} + 1 \\
&\approx 1, \quad \text{( } 1 < M \ll H' W', \; C_{\text{in}} \approx C_{\text{out}})
\end{aligned}
\end{equation}
Thus, compared to the standard convolution, the dynamic
convolution has about  \( M\)× parameters with negligible extra
FLOPs.
\subsubsection{Dual Dynamic Convolution Layer}
To improve the model’s adaptability to non-uniform visual patterns induced by varying environmental conditions, particularly in foggy scenarios, we introduce the Dual Dynamic Convolution Layer. This layer is designed to enhance the feature extraction process in complex and dynamic environments, replacing the standard double convolutional layers typically employed in architectures like U-Net with dynamic convolutional counterparts.

In traditional U-Net encoding stages, feature extraction is achieved through two consecutive, fixed convolutional layers. These layers progressively abstract the input features, but their fixed nature limits the model's ability to adapt to the complexities of varying environmental conditions. By incorporating dynamic convolutions, the Dual Dynamic Convolution Layer enables the network to adjust its convolution kernels based on the input characteristics. This dynamic modulation of convolution kernels provides a significant advantage in challenging conditions, such as fog, where visibility is reduced, and fine details become increasingly difficult to capture.

Formally, the output \( Y_1 \) from the first dynamic convolution is computed as:
\begin{equation}
\label{first_dyconv}
Y_1 = \sum_{k=1}^{K} \alpha^{(k)} \cdot (X * W^{(k)}),
\end{equation}
where \( X \) represents the input feature map, and \(\alpha^{(k)}\) are the input-dependent attention coefficients derived using an MLP module as described in Eq. \eqref{alpha}.

The output \( Y_2 \) from the second dynamic convolution can be expressed as:
\begin{equation}
Y_2 = \sum_{k=1}^{K} \beta^{(k)} \cdot (Y_1 * V^{(k)}),
\end{equation}
where \( V^{(k)} \) are the convolutional kernels of the second layer and \(\beta^{(k)}\) are attention coefficients generated based on \( Y \).

The dynamic convolution process enables iterative refinement of feature extraction, enabling the model to adaptively focus on relevant features under varying conditions. The Dual Dynamic Convolution Layer further enhances robustness in visually degraded environments, such as fog, where clarity and contrast are often reduced. By dynamically modulating convolution weights, the model preserves critical information, ensuring accurate detection of subtle features, such as road ponding boundaries, even under obfuscating conditions. Unlike traditional kernel-fixed convolutional layers, which rely on two consecutive fixed convolutions, dynamic convolutions provide greater flexibility by adjusting the kernels, improving feature extraction. Despite the increase in parameters, the model maintains computational efficiency, with a minimal rise in FLOPs, as demonstrated by the complexity analysis. This makes it suitable for applications such as road safety early-warning systems, where both accuracy and processing speed are critical.

\subsection{Wavelet-Based Synergy of Frequency and Spatial Information}

Detecting ponding in foggy environments presents significant challenges due to reduced visibility, light scattering, and the reflective properties of water surfaces. These factors often diminish scene clarity and detail, complicating the accurate identification of waterlogged areas using conventional image processing techniques. To address these challenges, we propose the Bidomain Information Synergy (BIS) module, an innovative framework designed to enhance detection performance under foggy conditions.

The BIS module comprises two synergistic pathways: the Frequency-Aware Information Processing Pathway and the Spatial-Enhanced Feature Learning Pathway. The frequency pathway processes low-frequency components to maintain global structural stability while leveraging high-frequency elements to enhance edge sharpness and capture intricate scene details. This dual pathway is particularly beneficial for accurately representing both the smooth and detailed aspects of water surfaces obscured by fog. In parallel, the spatial pathway emphasizes object localization and contextual relationships, contributing to the precise detection of waterlogged areas and their interactions with surrounding features. By integrating these pathways, the BIS module balances global coherence with local detail, enhancing feature representation and robustness in complex foggy environments.

\subsubsection{Frequency-Aware Information Processing Pathway}

\begin{figure*}[htbp]
    \centering    \includegraphics[width=1\textwidth]{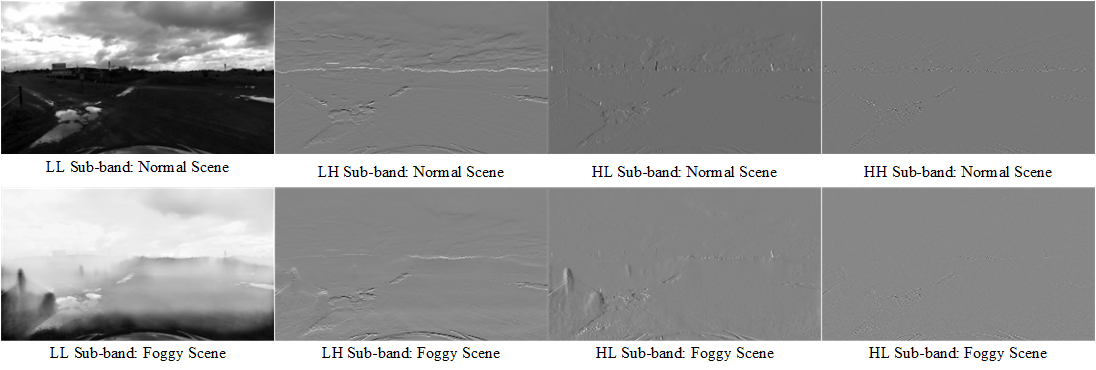}
    \caption{Wavelet Sub-band Analysis of Puddle-1000 Images and Synthesized Foggy-Puddle Images.}%The top row shows the four wavelet sub-bands (LL, LH, HL, HH) of the original Puddle-1000 images, highlighting clear edges and texture details in normal scenes. The bottom row presents the corresponding sub-bands of the synthesized foggy images, where significant degradation in high-frequency sub-bands (LH, HL, HH) can be observed, demonstrating the adverse impact of fog on texture and edge clarity.
    \label{fig:subband}
\end{figure*}

Foggy conditions pose significant challenges for ponding detection due to reduced visibility and the degradation of edge details, which are crucial for distinguishing ponded areas. These effects lead to a diminished clarity of the scene, particularly in the high-frequency components that capture fine details. To better understand how fog impacts both the frequency and spatial components of ponding detection, we apply the Discrete Wavelet Transform (DWT) to decompose the image into four sub-bands (See Fig. \ref{fig:subband}) : \( F_{\text{LL}} \), \( F_{\text{LH}} \), \( F_{\text{HL}} \), and \( F_{\text{HH}} \), which correspond to low- and high-frequency features. The LL sub-band primarily encodes the global structural context and remains relatively unaffected by fog, as evidenced by the minimal changes between normal and foggy scenes. In contrast, the high-frequency sub-bands (LH, HL, HH), which capture essential edge and texture information, exhibit substantial attenuation due to the fog's impact, leading to a loss of detail, particularly in areas where ponding is present.

The Frequency-Aware Information Processing Pathway aims to mitigate fog interference while preserving critical scene details to improve the detection of ponding areas. As illustrated in Fig.~\ref{fig:MIA} (b), this pathway consists of three main components: the Feature Mixing Block (FMBlock) \cite{fmblock, engineeringFre}, Discrete Wavelet Transform (DWT), and Inverse DWT (IDWT) \cite{dwt}. Given an input feature map \( F \in \mathbb{R}^{H \times W \times D} \), the DWT decomposes it into four sub-bands: \( F_{\text{LL}} \), \( F_{\text{LH}} \), \( F_{\text{HL}} \), and \( F_{\text{HH}} \), each of size \( \mathbb{R}^{\frac{H}{2} \times \frac{W}{2} \times D} \). The sub-band \( F_{\text{LL}} \) represents the low-frequency (approximation) component that preserves global structural context, while \( F_{\text{LH}} \), \( F_{\text{HL}} \), and \( F_{\text{HH}} \) capture high-frequency details such as edges and textures, essential for detailed scene representation \cite{wavedh}.

The DWT utilizes low-pass and high-pass filters defined as:
\begin{equation}
    l = \frac{1}{\sqrt{2}} \begin{bmatrix} 1 & 1 \end{bmatrix}, \quad
    h = \frac{1}{\sqrt{2}} \begin{bmatrix} -1 & 1 \end{bmatrix}.
\end{equation}
These filters extract both smooth (low-frequency) and detailed (high-frequency) components, ensuring the retention of essential scene elements. The sub-bands are computed as:
\begin{align}
    F_{\text{LL}}(u, v) &= \sum_{m=0}^{H-1} \sum_{n=0}^{W-1} F(m, n) \cdot l(m - 2u) \cdot l(n - 2v), \\
    F_{\text{LH}}(u, v) &= \sum_{m=0}^{H-1} \sum_{n=0}^{W-1} F(m, n) \cdot l(m - 2u) \cdot h(n - 2v), \\
    F_{\text{HL}}(u, v) &= \sum_{m=0}^{H-1} \sum_{n=0}^{W-1} F(m, n) \cdot h(m - 2u) \cdot l(n - 2v), \\
    F_{\text{HH}}(u, v) &= \sum_{m=0}^{H-1} \sum_{n=0}^{W-1} F(m, n) \cdot h(m - 2u) \cdot h(n - 2v),
\end{align}
where \( m \) and \( n \) are the row and column indices of \( F \), and \( u \) and \( v \) are the row and column indices of the sub-bands. The downsampling by a factor of 2 reduces the resolution of the sub-bands while retaining critical information.
\begin{figure}[htbp]
    \centering    \includegraphics[width=0.5\textwidth]{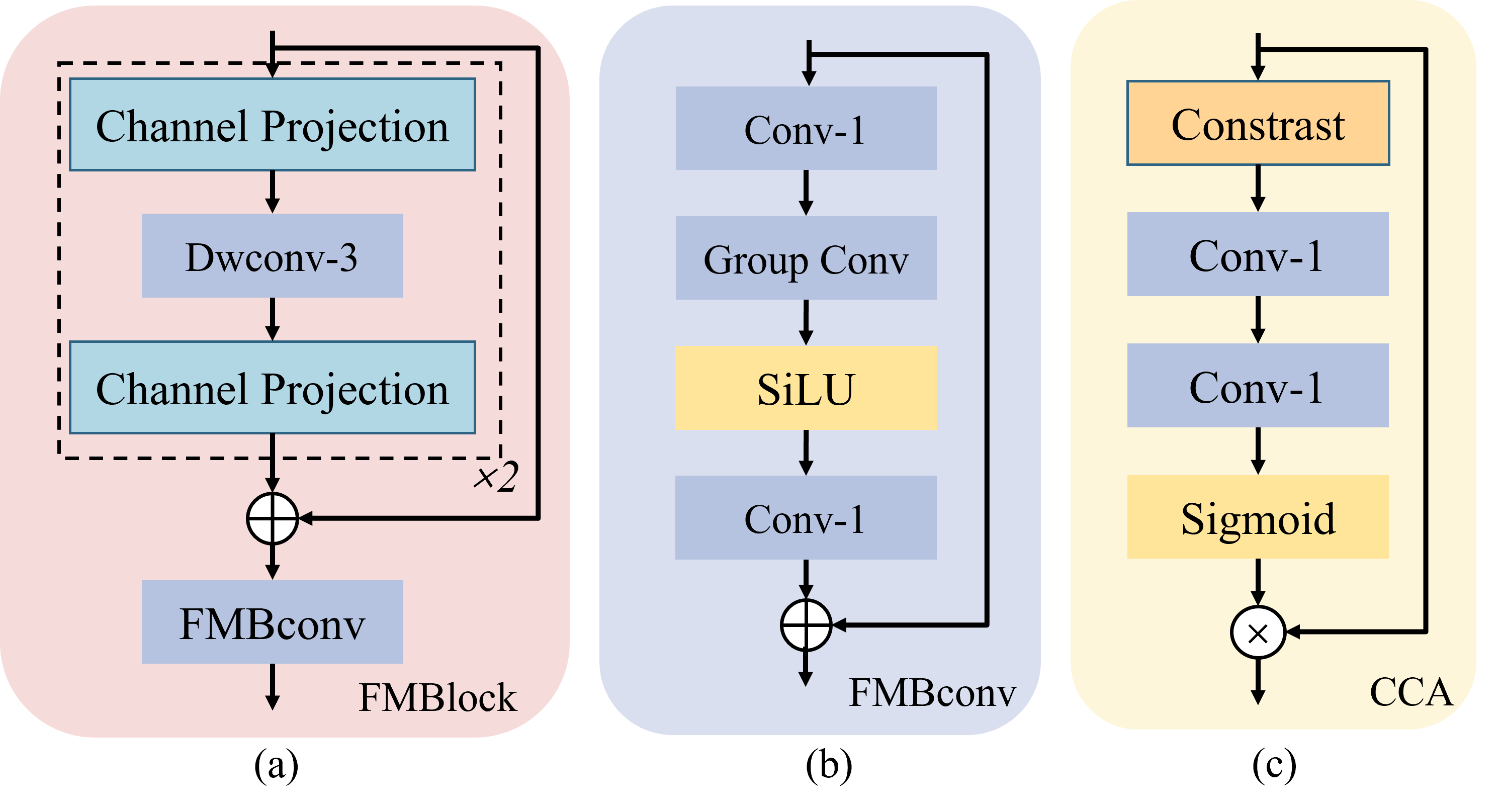}
    \caption{(a) The architecture of Feature Mixing Block (FMBlock). (b) The architecture of Feature Mixing Block Conv (FMBconv). (c) The architecture of Contrast-
Aware Channel Attention (CCA).}
    \label{fig:zujian}
\end{figure}
Traditional wavelet-based methods often process low- and high-frequency components uniformly, overlooking their distinct characteristics. This can hinder computational efficiency and limit the ability to address fog interference effectively. Our approach processes the low-frequency sub-band \( F_{\text{LL}} \) using the FMBlock, as shown in Fig. \ref{fig:zujian} (a) which enhances feature representation through depth-wise convolutions with large kernels to capture extensive contextual information. The FMBlock also incorporates channel splitting and shuffling mechanisms for effective feature interaction, enriching low-frequency representations while maintaining global structural coherence:
\begin{equation}
    \hat{F}_{\text{LL}} = \text{FMBlock}(F_{\text{LL}}),
\end{equation}

The enhanced low-frequency sub-band \( \hat{F}_{\text{LL}} \) and the high-frequency sub-bands are integrated via IDWT to reconstruct the feature map:
\begin{equation}
    \hat{F} = \text{IDWT}(\hat{F}_{\text{LL}}, F_{\text{LH}}, F_{\text{HL}}, F_{\text{HH}}),
\end{equation}
where \( \hat{F} \in \mathbb{R}^{H \times W \times D} \) combines both global and local information. Finally, a residual connection integrates the original input \( F \) with \( \hat{F} \), preserving the original information while incorporating enriched features:
\begin{equation}
    F_{\text{out, frequency}} = F + \hat{F},
\end{equation}
ensuring robust and adaptive detection in challenging conditions such as foggy environments.

\subsubsection{Spatial-Enhanced Feature Learning Pathway}

The Spatial-Enhanced Feature Learning Pathway enriches spatial feature representation through a sequence of specialized operations, including depth-wise convolutions, activation functions, and normalization layers. This design enables the extraction and refinement of local spatial features, providing essential details that complement the global structural insights obtained from the frequency pathway.

As depicted in Fig.~\ref{fig:over_architecture} (b), the process begins with a depth-wise convolutional to efficiently capture localized spatial patterns. A ReLU activation function follows, introducing non-linearity and enabling the network to learn complex, fine-grained representations. Batch normalization (\textit{BatchNorm}) is subsequently applied to stabilize feature distributions, promote training convergence, and enhance overall robustness. A second depth-wise convolutional layer further refines these spatial features, augmenting the pathway’s ability to delineate object boundaries and intricate spatial relationships:
\begin{equation}
    F' = \mathcal{D}_2\left( \mathcal{B}\left( \phi\left( \mathcal{D}_1(F) \right) \right) \right),
\end{equation}
where \( \mathcal{D}_1 \) and \( \mathcal{D}_2 \) represent the depth-wise convolution operations, \( \phi(\cdot) \) denotes the ReLU activation function, and \( \mathcal{B}(\cdot) \) refers to batch normalization.

To integrate the enriched spatial features with the original input seamlessly, a residual connection is incorporated:
\begin{equation}
    F_{\text{out, spatial}} = F' + F,
\end{equation}
ensuring that the model retains the initial information while embedding enhanced spatial details. This balanced approach strengthens the overall feature representation, equipping the model for precise ponding detection under challenging conditions such as foggy environments, where maintaining both local and global information is critical.

After processing through both the Frequency-Aware Information Processing Pathway and the Spatial-Enhanced Feature Learning Pathway, the outputs are concatenated and refined using a depth-wise separable convolution to achieve cohesive feature integration:
\begin{equation}
    F_{\text{final}} = \mathcal{D}_3\left( \text{Concat}\left( F_{\text{out, frequency}}, F_{\text{out, spatial}} \right) \right),
\end{equation}
where \( \text{Concat}(\cdot) \) denotes the concatenation operation, \( \mathcal{D}_3(\cdot) \) refers to the depth-wise separable convolution, and \( F_{\text{out, frequency}} \) and \( F_{\text{out, spatial}} \) represent the outputs from the frequency and spatial pathways, respectively. This synergistic operation fuses information from both the frequency and spatial domains, enhancing the model's capability to capture comprehensive feature representations. The result is a feature map \( F_{\text{final}} \) that effectively balances global structural context with local fine details, ensuring robust performance in complex and challenging foggy environments.

\subsection{Aggregation-Broadcast-Coupling Mechanism}

To address the challenge of detecting ponding at multiple spatial scales, we propose the Aggregation-Broadcast-Coupling (ABC) mechanism. This mechanism comprises two key components: the Multi-Scale Information Aggregation (MIA) module and the Adaptive Attention Coupling Gate (AACG) module. The ABC mechanism aggregates features across different levels and broadcasts global contextual information to each level, facilitating the effective integration of features from multiple ponding scales while selectively enhancing crucial detection cues. In the following sections, we provide a comprehensive description of these two modules.

\subsubsection{Multi-Scale Information Aggregation Module}
\begin{figure*}[htbp]
    \centering
    \includegraphics[width=1\textwidth]{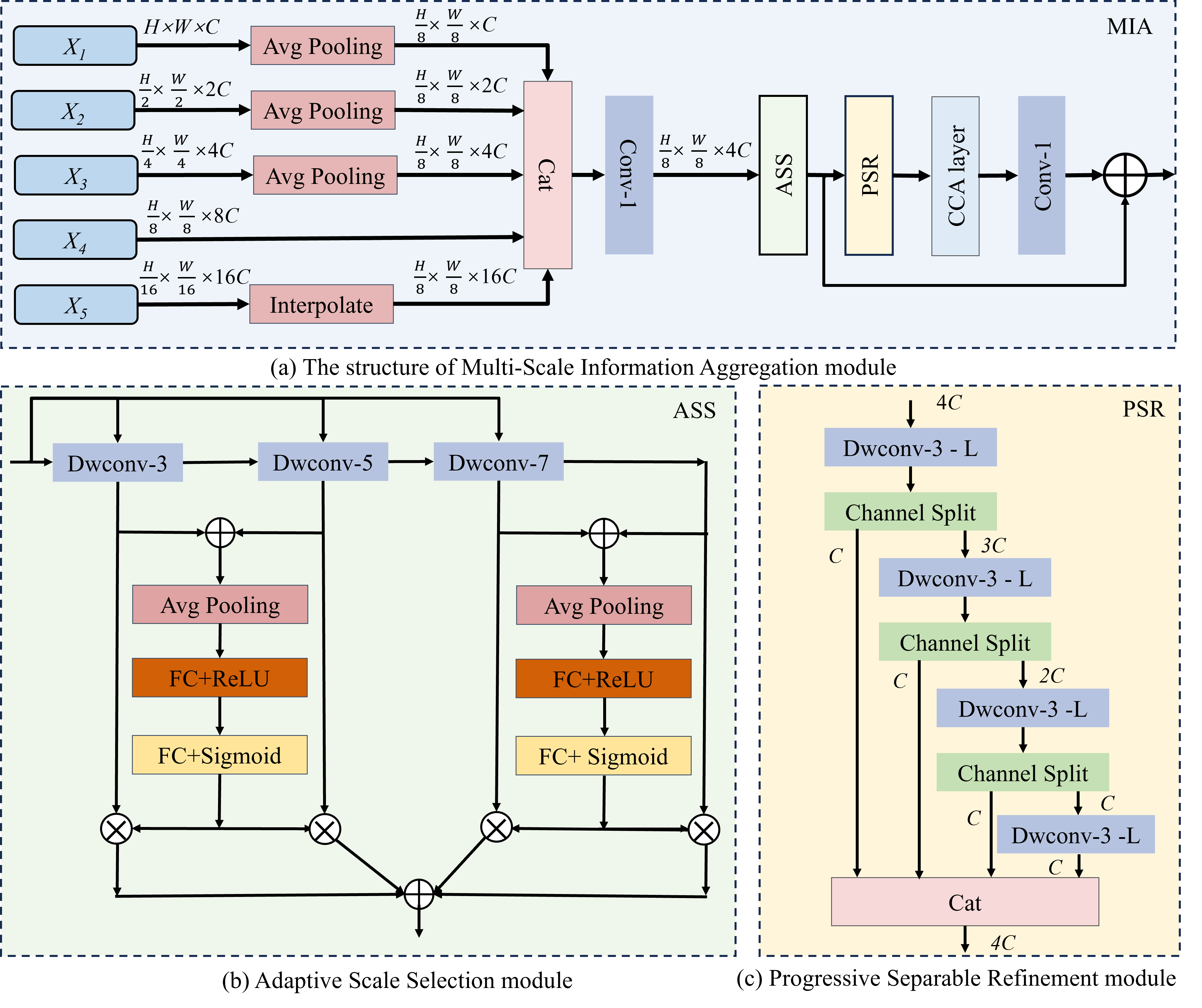}  
    \caption{The Architecture of Multi-Scale Information Aggregation module. (a) illustrates the oversall computational process of MIA module. (b) details the computational specifics of the Adaptive Scale Selection (ASS). (c) shows the Progressive Separable Refinement module.}
    \label{fig:MIA} 
\end{figure*}
The Multi-Scale Information Aggregation (MIA) module, as shown in Fig.\ref{fig:MIA} (a), is a core component of the Aggregation-Broadcast-Coupling (ABC) mechanism, designed to tackle the challenges of detecting ponding on foggy roads. In such environments, ponding areas are affected by factors such as reduced visibility, intricate road structures, and uneven water distribution. To effectively detect ponding under these challenging conditions, it is crucial for the system to capture features at multiple spatial scales, allowing it to distinguish between both broad contextual information and fine-grained details.

The MIA module integrates two key processes: Adaptive Scale Selection (ASS), as shown in Fig.\ref{fig:MIA} (b), and Progressive Separable Refinement (PSR), as shown in Fig.\ref{fig:MIA} (c). Together, these components synergistically capture, refine, and enhance multi-scale features, producing contextually aware feature representations that improve detection accuracy.

The first step in the process involves concatenating feature maps from all levels, denoted as \( \{F_1, F_2, F_3, F_4, F_5\} \). Following this, adaptive average pooling (AAP) or bilinear interpolation is applied to unify the number of channels across these feature maps to match that of \( F_4 \). A subsequent 1x1 convolution is then performed to reduce the dimensionality of the concatenated feature maps to \( F_4 \)'s channel size, thereby mitigating computational overhead and reducing resource consumption. This step can be formally expressed as:

\begin{equation}
    \mathbf{F}_{\text{all}} = W_1 \left( \text{AAP} \left( \text{Concat}(F_1, F_2, F_3, F_4, F_5), C_4 \right) \right)
\end{equation}

where \( W_1 \) represents the weights of the 1x1 convolution and \( C_4 \) denotes the target channel size for unification.

The second step is Adaptive Scale Selection (ASS), which is essential for detecting ponding across a range of spatial extents. Ponding areas can vary dramatically in size, from small puddles to large, extensive waterlogged regions. To address this scale variation, ASS uses convolutions with different receptive field sizes—3x3, 5x5, and 7x7 kernels—to generate three distinct feature maps. Additionally, a sequence of convolutions (3x3, 5x5, and 7x7) is applied to generate a fourth feature map that aggregates information across all scales. The final aggregated feature map \( \mathbf{F}_{\text{ASS}} \) is computed as a weighted sum of these four feature maps, where the weights \( w_k \) are dynamically learned to prioritize the most relevant scales for the detection task:

\begin{equation}
    \mathbf{F}_{\text{ASS}} = \sum_{k \in \{k_1, k_2, k_3, \text{chain}\}} w_k \cdot F_k(X),
\end{equation}

Here, \( F_{k_1}(X), F_{k_2}(X), F_{k_3}(X) \) represent the feature maps generated by the 3x3, 5x5, and 7x7 convolutions, and \( F_{\text{chain}}(X) \) is the feature map produced by the sequential application of these kernels. The dynamic learning of the weights \( w_k \) ensures that the network focuses on the most pertinent scales, enhancing its ability to detect ponding over different spatial extents.

Once the multi-scale features are aggregated via ASS, the process advances to Progressive Separable Refinement (PSR). This step is critical for refining the aggregated features, progressively improving spatial details and boundary delineation—particularly important for foggy conditions where blurred boundaries can complicate detection. Fog-induced blur may obscure the boundaries of waterlogged areas, making accurate detection more difficult. To mitigate this, PSR employs a series of depthwise separable convolutions to progressively refine the aggregated feature map \( \mathbf{F}_{\text{ASS}} \), enhancing spatial precision while preserving the integrity of the multi-scale features.

At each refinement step, PSR splits a subset of channels, referred to as "distilled channels," which are iteratively refined to capture fine spatial details. The remaining channels is preserved. This iterative refinement process can be formalized as follows:

\begin{equation} 
    \mathbf{F}_{i, \text{dis}}, \mathbf{F}_{i, \text{rem}} = 
    \begin{cases}
    \text{Split}_1 \left( \mathcal{DL}_1 (\mathbf{F}_{\text{in}}) \right), & i = 1 \\
    \text{Split}_i \left( \mathcal{DL}_i (\mathbf{F}_{i-1, \text{rem}}) \right), & i = 2, \dots, N-1 \\
    \mathcal{DL}_N (\mathbf{F}_{N-1, \text{rem}}), & i = N
    \end{cases}
\end{equation}

where \( \mathbf{F}_{i, \text{dis}} \) represents the distilled channels at each step, and \( \mathbf{F}_{i, \text{rem}} \) denotes the remaining channels processed in subsequent layers. Here, \( \mathcal{DL}_i \) refers to the depthwise separable convolution with a Leaky ReLU activation at the \( i \)-th step, while \( \text{Split}_i \) denotes the channel separation operation.

After the final refinement step, the distilled channels \( \mathbf{F}_{i, \text{dis}} \) from all iterations are concatenated to form the final distilled feature map:

\begin{equation}
    \mathbf{F}_{\text{distilled}} = \text{Concat} \left( \mathbf{F}_{1, \text{dis}}, \mathbf{F}_{2, \text{dis}}, \dots, \mathbf{F}_{N, \text{dis}} \right)
\end{equation}

The output \( \mathbf{F}_{\text{distilled}} \) consolidates these iterative enhancements, ensuring that both fine spatial details and boundaries are preserved with high clarity, even in challenging foggy conditions.

In addition to refining spatial details, PSR and a Contrast-Aware Channel Attention (CCA) mechanism. This mechanism enhances the model's ability to prioritize channels based on contrast sensitivity, which is crucial in foggy environments where the contrast between waterlogged areas and the surrounding surfaces may be weak or distorted. By focusing on channels that exhibit higher contrast and suppressing less relevant ones, CCA enables the model to concentrate on the most discriminative features for accurate ponding detection.The details of CCA as shown in Fig. \ref{fig:zujian} (c). 

The CCA mechanism computes attention weights based on the global contrast within each channel. It captures both the mean and variance of the feature map to reflect contrast variations. These contrast values are processed through learnable weight matrices and non-linear activations to generate the attention map \( \zeta \), which selectively amplifies channels with high contrast:

\begin{equation}
    \zeta = \sigma \left( W_3 \cdot \phi \left( W_2 \cdot \text{Contrast}(\mathbf{F}_{\text{distilled}}) \right) \right)
\end{equation}

where \( \text{Contrast}(\mathbf{F}_{\text{distilled}}) \) computes the mean and variance for each channel in \( \mathbf{F}_{\text{distilled}} \), capturing the contrast profile. The attention map \( \zeta \) is then applied to the feature map through element-wise multiplication:

\begin{equation}
    \mathbf{F}_{\text{CCA}} = \zeta \odot \mathbf{F}_{\text{distilled}}
\end{equation}

By selectively amplifying channels sensitive to contrast variations, CCA helps the model focus on subtle features critical for ponding detection, especially in low-contrast, foggy conditions. This attention mechanism significantly enhances the model's robustness and accuracy in detecting ponding areas, even when visual degradation obscures critical details.

\subsubsection{Adaptive Attention Coupling Gate} 

The Adaptive Attention Coupling Gate (AACG) module is designed to enhance feature fusion in the skip connections by adaptively controlling the information flow between different feature sources, specifically the Multi-Scale Information Aggregation (MIA) module and the each encoding stages. This module plays a crucial role in ensuring that only contextually relevant information is retained and amplified, which is particularly beneficial for complex tasks such as detecting ponding in foggy environments, where certain spatial features need emphasis.

In the AACG module, fusion starts with a Multi-Head Cross Attention mechanism, which integrates feature maps from the MIA and each encoding stage. Let \( F_{\text{MIA}} \) and \( F_{\text{encoder}} \) be the input feature maps. The Query (\( Q \)), Key (\( K \)), and Value (\( V \)) matrices for each attention head \( i \) are computed as:

\begin{equation}
Q_i = W_Q^i F_{\text{MIA}}, \quad K_i = W_K^i F_{\text{encoder}}, \quad V_i = W_V^i F_{\text{encoder}}
\end{equation}

The Multi-Head Self-Attention (MHSA) is then computed by concatenating the attention heads and applying a final transformation:

\begin{equation}
\text{MHSA}(Q, K, V) = \text{Concat}\left( \text{head}_1, \text{head}_2, \dots, \text{head}_h \right)
\end{equation}

\begin{equation}
\text{head}_i = \text{Softmax}\left( \frac{Q_i K_i^T}{\sqrt{d_k}} \right) V_i
\end{equation}

Here, \( d_k \) is the dimensionality of the key vector for each attention head. The final fused representation selectively integrates the most relevant contextual information from both the MIA and encoding stage feature maps, capturing essential spatial details.

To further refine this fused output, AACG applies Layer Normalization to \( F_{\text{fused}} \), stabilizing the feature distribution and improving training robustness. This normalized feature map is then passed through a Sigmoid activation, producing a gating signal \( \lambda \) that ranges from 0 to 1. This signal functions as an adaptive gate, modulating the relative importance of each feature channel based on its relevance. The Sigmoid-activated gating signal is calculated as:
\begin{equation}
\lambda = \sigma(\text{LayerNorm}(F_{\text{fused}}))
\end{equation}
where \( \sigma \) represents the sigmoid function.

The final stage in the AACG module is the application of this gating signal to the encoder feature map. Through element-wise multiplication, the gating signal selectively amplifies or suppresses features in \( F_{\text{fused}} \), producing the final gated output:
\begin{equation}
F_{\text{gated}} = \lambda \odot F_{\text{encoder}}
\end{equation}
where \( \odot \) denotes element-wise multiplication.

To preserve the original information from MIA while incorporating the adaptively gated features, a residual connection is applied between the MIA feature map and the gated output. The final output of the AACG module, \( F_{\text{output}} \), is obtained by adding \( F_{\text{MIA}} \) back to \( F_{\text{gated}} \):
\begin{equation}
F_{\text{output}} = F_{\text{gated}} + F_{\text{MIA}}
\end{equation}

This residual connection helps maintain essential spatial information from the MIA while allowing the network to adaptively enhance contextually relevant details. The gated output ensures that only the most contextually relevant features are retained and propagated through the skip connection, enhancing the network's ability to maintain critical spatial and contextual details.

In summary, the AACG module serves as an adaptive feature gate, integrating the most relevant features from both the MIA and each encoding stages feature maps through a combination of Cross Attention, Layer Normalization, and Sigmoid-based gating, followed by a residual connection. This fusion mechanism enables the network to focus selectively on the most important features, ensuring robust and accurate detection of ponding in complex scenarios, such as foggy environments where subtle features must be preserved to improve detection performance.

\section{Experiments}
In this section, we begin by introducing the datasets: Puddle-1000, Foggy-Puddle, and Foggy Low-light Puddle. Next, we present the experimental setup and the evaluation metrics employed. To assess the impact of individual components on the overall network performance, we conduct ablation studies. Following this, we perform a comparative analysis against state-of-the-art methods on these datasets. Lastly, we evaluate the computational efficiency of the proposed network on an edge computing platform and provide a detailed analysis of the results.
\subsection{Datasets, Setting, and Evaluation Metrics}
\subsubsection{Datasets}
We evaluate the proposed method using three datasets: Puddle-1000, Foggy-Puddle, and Foggy Low-light Puddle, providing a comprehensive assessment of its effectiveness and robustness in challenging conditions.
\paragraph{Puddle-1000} The Puddle-1000 dataset, introduced by Han et al. \cite{waterhazard}, is one of the few publicly available benchmarks specifically designed for road ponding detection under normal, clear-weather conditions. Comprising 985 manually labeled RGB images captured with a ZED camera, this dataset provides a critical resource for evaluating algorithm performance in structured and unstructured environments. It is divided into two subsets tailored to different driving scenarios: the on-road (ONR) subset with 357 structured ponding images and the off-road (OFR) subset containing 628 unstructured images, both at a resolution of 640×360. This segmentation allows for targeted analysis of algorithm robustness across diverse road conditions.

\paragraph{Foggy-Puddle}Building on the Puddle-1000 dataset, the Foggy-Puddle dataset was synthesized using an atmospheric scattering model, as shown in Eq. \eqref{daqi}, to simulate foggy conditions \cite{agsenet}. This dataset leverages the atmospheric scattering equation:

\begin{equation}
\label{daqi}
I(x) = J(x) \cdot t(x) + A \cdot (1 - t(x)),
\end{equation}

where \( I(x) \) represents the foggy image, \(J(x) \) is the clear-weather image, \(t(x) \) denotes the medium transmission map, and \( A \) indicates the global atmospheric light. The transmission map \(t(x) \) is typically computed using the following formula: 
\begin{equation}
\label{tx}
t(x) = e^{-\kappa d(x)},
\end{equation}

where $ \kappa $ is the scattering coefficient that controls the density of the fog, and \(d(x) \) is the depth of the scene at pixel location \(x \). The Eq. \eqref{tx} the exponential decay of light as it passes through the foggy medium, with larger values of $ \beta $ indicating denser fog and greater attenuation of light.
The primary goal of the Foggy-Puddle dataset is to introduce a challenging benchmark that enables the assessment of road ponding detection methods under adverse weather conditions, specifically fog-induced visibility degradation. By transforming the clear-weather images from Puddle-1000 into foggy scenarios, this dataset facilitates comprehensive evaluation of algorithm resilience to atmospheric distortions.

\paragraph{Foggy Low-light Puddle} Extending the capabilities of road ponding detection to low-light and reduced-visibility conditions, we developed the Foggy Low-light Puddle dataset as a specialized benchmark for testing algorithms in these challenging environments. Building on the Night-Puddle dataset, which comprises 500 images captured in low-light and near-darkness conditions from urban, suburban, and campus roads across ten cities in China. We employed the Monodepth2 algorithm \cite{monodepth2} to generate precise depth maps from the Night-Puddle images, using these maps to guide the application of an atmospheric scattering model that simulates realistic fog effects. The depth information was key to this process, allowing us to adjust the fog density based on the distance of objects in the scene, creating a natural variation where fog appears denser in distant areas while preserving details in closer regions. This approach not only accurately captures the combined challenges of foggy, low-light conditions but also provides a solid foundation for evaluating algorithm performance in these compounded scenarios. By integrating both fog and low-light elements into a single benchmark, the Foggy Low-light Puddle dataset addresses a significant gap in road ponding research, offering a crucial resource for developing and validating algorithms in adverse environments. This dataset plays a key role in advancing research on multi-condition road safety, with a specific focus on the complexities of detecting road ponding under foggy, low-light conditions.
\subsubsection{Setting}
\paragraph{Data Preprocessing}In our experiments, all models are trained from scratch without using any pre-trained weights. During preprocessing, all images and their corresponding target masks are uniformly resized to 256×256 pixels. To stabilize training, the pixel values are normalized to the [0, 1] range. 

\paragraph{Training Procedure}We initiate training with an AdamW optimizer, using an initial learning rate of $1 \times 10^{-4}$. The learning rate follows a cosine annealing schedule, which periodically reduces the learning rate to allow the model to escape potential local minima and improve convergence. We set the batch size to 8. Training proceeds for 100 epochs or until early stopping is triggered based on validation loss.

\paragraph{Computational Environment} All experiments are conducted using PyTorch 2.1.1, and model evaluations are performed on a single NVIDIA GeForce RTX 4090 GPU, running CUDA 11.8, Python 3.10.13, and cuDNN 8.7.0. 
\subsubsection{Evaluation Metrics}
To quantitatively evaluate the performance of our segmentation model in detecting road ponding, we employ four widely recognized metrics: Intersection over Union (IoU), F1 Score, Mean Intersection over Union (MIoU), and Mean Pixel Accuracy (MPA). These metrics were selected to provide a comprehensive assessment of the model’s effectiveness and robustness.

\paragraph{Intersection over Union (IoU)}
IoU is a commonly used metric for evaluating the overlap between the predicted segmentation and the ground truth. It is calculated as the ratio of the intersection (true positives) to the union of the predicted and ground truth regions (true positives, false positives, and false negatives). Mathematically, IoU is expressed as:
\begin{equation}
\label{iou}
\text{IoU} = \frac{TP}{TP + FP + FN},
\end{equation}
where \( TP \), \( FP \), and \( FN \) represent the true positives, false positives, and false negatives, respectively.

\paragraph{F1 Score}
The F1 Score is the harmonic mean of precision and recall, balancing both metrics to provide a singular measure of accuracy. It is particularly useful in scenarios where the class distribution is imbalanced. The F1 Score is defined as:
\begin{equation}
\label{f1}
\text{F1} = \frac{2 \times TP}{2 \times TP + FP + FN},
\end{equation}

\paragraph{Mean Intersection over Union (MIoU)}
MIoU extends the IoU metric to multi-class segmentation tasks by averaging the IoU across all classes. It is calculated as:
\begin{equation}
\label{miou}
\text{MIoU} = \frac{1}{C} \sum_{i=1}^{C} \frac{TP_i}{TP_i + FP_i + FN_i},
\end{equation}
where \( C \) denotes the total number of classes, and \( TP_i \), \( FP_i \), and \( FN_i \) represent the true positives, false positives, and false negatives for each class \( i \).

\paragraph{Mean Pixel Accuracy (MPA)}
MPA measures the average accuracy per pixel across all classes. It considers both true positives and true negatives, providing insight into pixel-wise classification performance. The formula for MPA is:
\begin{equation}
\label{mpa}
\text{MPA} = \frac{1}{C} \sum_{i=1}^{C} \frac{TP_i + TN_i}{TP_i + TN_i + FP_i + FN_i},
\end{equation}
where \( TN_i \) represents the true negatives for class \( i \), and the other terms are as previously defined. MPA ensures that both foreground (water) and background (non-water) regions are evaluated, offering a balanced perspective on pixel-wise classification accuracy.

\subsection{Ablation Studies}
In our comprehensive ablation experiments on the Foggy-Puddle dataset, we delve into the individual and collective contributions of three key components of the proposed model: the Bidomain Information Synergy (BIS) module, the Multi-scale Information Aggregation (MIA) module, and the Adaptive Attention Coupling Gate (AACG) module. Our detailed analysis, summarized in Table \ref{tab:ablation_study}, quantifies the impact of each component on puddle detection performance under foggy conditions. In addition to these core elements, we provide an in-depth evaluation of the convolution types within the Dual Dynamic Convolution (DDC) layer, as detailed in Table \ref{tab:conv_ablation_study}. Furthermore, we examine the influence of the number of dynamic experts in the DDC layer on the overall efficacy and efficiency of ABCDWaveNet, with the results presented in Table \ref{tab:expert_ablation_study}.

%\begin{comment}
\begin{table*}[ht]
\centering
\begin{threeparttable}
\caption{Ablation study on the proposed components of ABCDWaveNet. \textcolor{red}{Red} values indicate a deterioration compared to the baseline configuration, while \textcolor{blue}{blue} values indicate an improvement. \textbf{Bold} values represent the best performance. w/o stands for without.}
\label{tab:ablation_study}
\begin{tabular}{lllllll}
\toprule
\textbf{Models}             & \textbf{IoU} & \textbf{F1 Score} & \textbf{MIoU} & \textbf{MPA} & \textbf{\#Param(M)} & \textbf{GFLOPs} \\ \midrule
\rowcolor{blue!8} ABCDWaveNet (Full Model)    & \textbf{84.59} & \textbf{92.22} & \textbf{91.65} & \textbf{95.74} & 47.51 & 38.16 \\ 
w/o BIS            &83.72 \textcolor{red}{(-0.87)}  &91.78 \textcolor{red}{(-0.44)}  &91.14 \textcolor{red}{(-0.51)}  &95.34 \textcolor{red}{(-0.40)} &37.33 \textcolor{blue}{(-10.18)}   & 28.62 \textcolor{blue}{(-9.54)}  \\
w/o ASS                   &84.16 \textcolor{red}{(-0.43)}   &92.01 \textcolor{red}{(-0.21)}  &91.40 \textcolor{red}{(-0.25)}   & 95.65 \textcolor{red}{(-0.09)}  & 47.50 \textcolor{blue}{(-0.01)}   & 37.71 \textcolor{blue}{(-0.45)}  \\
w/o PSR                  &84.25 \textcolor{red}{(-0.34)}   &92.05 \textcolor{red}{(-0.17)}   &91.47 \textcolor{red}{(-0.18)}  &95.70 \textcolor{red}{(-0.04)}   &47.24 \textcolor{blue}{(-0.27)}   &37.90 \textcolor{blue}{(-0.26)}  \\
w/o AACG               &84.04 \textcolor{red}{(-0.55)}  &91.99 \textcolor{red}{(-0.23)}  &91.37 \textcolor{red}{(-0.28)}  &95.59 \textcolor{red}{(-0.15)}    &47.01 \textcolor{blue}{(-0.50)}   &26.32 \textcolor{blue}{(-11.84)}  \\
w/o BIS + AACG         & 83.67 \textcolor{red}{(-0.92)}  & 91.76 \textcolor{red}{(-0.46)}  & 91.11 \textcolor{red}{(-0.54)}  & 95.30 \textcolor{red}{(-0.44)}     &36.84 \textcolor{blue}{(-10.67)}   &16.78 \textcolor{blue}{(-21.38)}   \\
w/o MIA + AACG           &83.86 \textcolor{red}{(-0.73)}   &91.86 \textcolor{red}{(-0.36)}  &91.23 \textcolor{red}{(-0.42)}  &95.36 \textcolor{red}{(-0.38)}  &45.69 \textcolor{blue}{(-1.82)}   &25.01 \textcolor{blue}{(-13.15)}  \\
w/o BIS + MIA + AACG     &83.10 \textcolor{red}{(-1.49)}  &91.47 \textcolor{red}{(-0.75)}  &90.71 \textcolor{red}{(-0.94)}  &95.27 \textcolor{red}{(-0.47)} &\textbf{35.52} \textcolor{blue}{(-11.99)}  &\textbf{15.47} \textcolor{blue}{(-22.69)}  \\ 
\bottomrule
\end{tabular}
\begin{tablenotes}
\small
\item Note: We evaluate the IoU, F1 Score, MIoU, and MPA metrics on the Foggy-Puddle dataset, and measure the parameter size and GFLOPs. BIS refers to Bidomain Information Synergy, ASS stands for Adaptive Scale Selection, PSR denotes Progressive Sparable Refinement, MIA refers to Multi-scale Information Aggregation, and AACG is the Adaptive Attention Coupling Gate. 
\end{tablenotes}
\end{threeparttable}
\end{table*}

\begin{figure*}[htbp]
    \centering
    \includegraphics[width=1\textwidth]{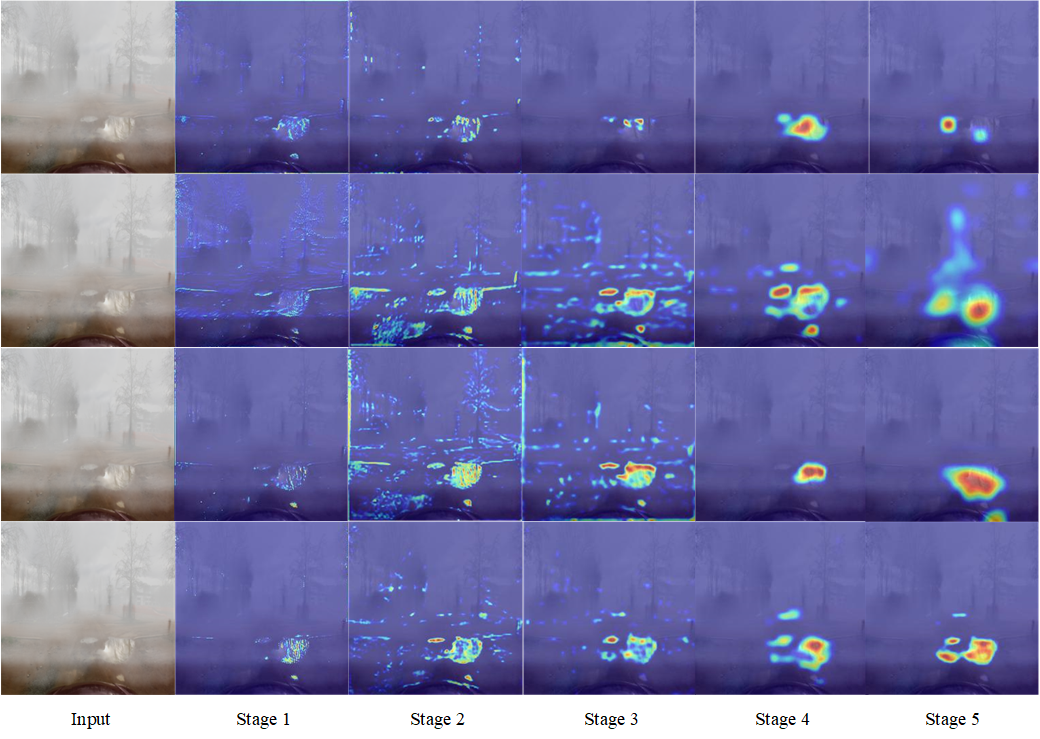}  
    \caption{Heatmaps of encoder activations across stages (Stage 1 to Stage 5), presented from left to right and top to bottom for the following configurations: baseline, baseline with the BIS module, baseline with the ABC mechanism, and baseline with both BIS and ABC (ours).}
    \label{fig:heatmap} 
\end{figure*}

\subsubsection{Analysis of Bidomain Information Synergy (BIS) module}The BIS module integrates a frequency-aware information pathway that processes low-frequency components to sustain a stable global structure, while high-frequency elements enhance sharpness and preserve intricate details. Meanwhile, the Spatial-Enhanced Feature Learning Pathway bolsters object localization and relational context across the scene. To quantify the importance of BIS, we conduct
experiments on a model variant devoid of the BIS module (w/o BIS). The exclusion of the BIS module led to a significant decline in model performance, with the IoU decreasing from 84.59\% to 83.72\%, a drop of 0.87\%, and the F1 Score falling from 92.22\% to 91.78\%, a reduction of 0.44\%. These performance metrics underscore the essential role of the BIS module in preserving critical details while effectively mitigating haze interference.
Furthermore, removing the BIS module reduced the number of parameters from 47.51 million to 37.33 million and decreased GFLOPs from 38.16 to 28.62, achieving computational efficiency. However, this efficiency gain came at the cost of significant performance reduction. Thus, the computational cost is worthwhile.
\subsubsection{Analysis of Aggregation-Broadcast-Coupling Mechanism} Our systematic analysis of the Aggregation-Broadcast-Coupling (ABC) mechanism, which is strategically formulated to effectively integrate multi-scale information. The ABC mechanism consists of a Multi-scale Information Aggregation (MIA) module and an Adaptive Attention Coupling Gate (AACG). The MIA module is designed to combine features from various each encoding stages in a coarse-to-fine manner, ensuring rich detail in the output through effective multi-scale feature fusion. To investigate MIA's functionality in progressively capturing multi-scale information, we conducted an ablation study by removing the Adaptive Scale Selection (ASS) module, which learns the structure and context at a coarse-grained level, and the Progressive Separable Refinement (PSR) module, which distills high-discriminative features. As a result, the IoU decreased by 0.43 and 0.34, respectively.

To further examine the contribution of AACG, we simplified the model by removing AACG and substituting the concatenation operation with element-wise addition, a simpler fusion method. As demonstrated in Table \ref{tab:ablation_study}, the full model achieves an IoU of 84.59. Removing the AACG module (w/o AACG) led to a marked decrease in IoU to 84.04, indicating a 0.55 reduction. When the entire ABC mechanism was excluded (w/o MIA+AACG), model performance significantly declined, with the IoU dropping by 0.73 to 83.86.

\begin{table*}[ht]
\centering
\caption{Ablation study for different configurations of experts in Dynamic Convolution.  \textcolor{red}{Red} values indicate a deterioration compared to the baseline configuration, while \textcolor{blue}{blue} values indicate an improvement. \textbf{Bold} values represent the best performance.}%Results are evaluated on the Foggy-Puddle dataset, reporting IoU, F1 Score, MIoU, MPA, parameters size, and GFLOPs.
\label{tab:expert_ablation_study}
\begin{tabular}{lllllll}
\toprule
\textbf{Num\_experts}  & \textbf{IoU} & \textbf{F1 Score} & \textbf{MIoU} & \textbf{MPA} & \textbf{\#Param(M)} & \textbf{GFLOPs} \\ 
\midrule
\rowcolor{blue!8} {[2, 2, 2, 4, 4]} & 84.59 & 92.22 & 91.65 & 95.64 & 47.51 & 38.16 \\ 
{[2, 2, 2, 2, 2]} & 84.05 \textcolor{red}{(-0.54)} & 91.95 \textcolor{red}{(-0.27)} & 91.33 \textcolor{red}{(-0.32)} & 95.55 \textcolor{red}{(-0.09)} & 30.99 \textcolor{blue}{(-16.52)} & 38.14 \textcolor{blue}{(-0.02)} \\ 
{[4, 4, 4, 2, 2]} & 84.21 \textcolor{red}{(-0.38)} & 92.04 \textcolor{red}{(-0.18)} & 91.44 \textcolor{red}{(-0.21)} & 95.66 \textcolor{blue}{(+0.02)} & 33.28 \textcolor{blue}{(-16.52)} & 38.14 \textcolor{blue}{(-0.02)} \\ 
{[4, 4, 4, 4, 4]} & \textbf{84.62} \textcolor{blue}{(+0.03)} & \textbf{92.28} \textcolor{blue}{(+0.06)} & \textbf{91.70} \textcolor{blue}{(+0.05)} & \textbf{95.68} \textcolor{blue}{(+0.04)} & 49.80 \textcolor{red}{(+2.29)} & 38.16 \textcolor{blue}{(+0.00)} \\ 
\bottomrule
\end{tabular}
\end{table*}
\subsubsection{Analysis of Different Configuration of Dynamic Experts}The number of dynamic experts directly impacts model complexity. We evaluated various configurations of experts across dual convolution layers as shown in Table \ref{tab:expert_ablation_study}. Initially, assigning two experts to the first three encoding stages and four experts to the subsequent two stages achieved optimal performance, striking a balance between accuracy and computational efficiency. In contrast, setting only two experts for all stages, although computationally beneficial, resulted in a significant drop in performance. This indicates that such a configuration may be less suitable for applications where accuracy is a priority.

We also explored a configuration with four experts in the earlier encoding stages and two experts in the deeper stages. This arrangement led to a 0.38 reduction in IoU. This is due to the complex features and rich contextual information in the deeper encoding stages, which require a higher representational capacity. With fewer experts in these stages, the model's ability to capture intricate details is limited. Additionally, uniformly increasing the number of experts across all stages to [4, 4, 4, 4, 4] yielded only negligible accuracy gains. While a higher number of experts theoretically enhances representational power, it also complicates the simultaneous optimization of multiple convolution kernels and attention mechanisms, increasing the risk of overfitting. This configuration also creates a significant computational burden, and the slight accuracy gains become negligible compared to the rising computational cost.

\begin{table*}[ht]
\centering
\caption{Ablation study on different convolution types in Dual Convolutional Layers. \textcolor{red}{Red} values indicate a deterioration compared to the baseline configuration, while \textcolor{blue}{blue} values indicate an improvement. \textbf{Bold} values represent the best performance.} %Results are evaluated on the Foggy-Puddle dataset, reporting IoU, F1 Score, MIoU, MPA, parameter size, and GFLOPs. 
\label{tab:conv_ablation_study}
\renewcommand\arraystretch{1}
\begin{tabular}{lllllll}
\toprule
\textbf{Convolution Type}  & \textbf{IoU} & \textbf{F1 Score} & \textbf{MIoU} & \textbf{MPA} & \textbf{\#Param(M)} & \textbf{GFLOPs} \\ 
\midrule
\rowcolor{blue!8} \textbf{Dynamic Convolution} & \textbf{84.59} & \textbf{92.22} & \textbf{91.65} & \textbf{95.74} & 47.51 & 38.16 \\ 
Standard Convolution &82.33 \textcolor{red}{(-2.26)} &90.94 \textcolor{red}{(-1.28)} &89.93 \textcolor{red}{(-1.72)} &95.16 \textcolor{red}{(-0.58)} &21.58 \textcolor{blue}{(-25.93)} &38.11 \textcolor{blue}{(-0.05)} \\ 
Dilated Convolution &79.49 \textcolor{red}{(-5.10)} &89.64 \textcolor{red}{(-2.58)} &88.57 \textcolor{red}{(-3.08)} &94.09 \textcolor{red}{(-1.65)} &21.58 \textcolor{blue}{(-25.93)} &38.11 \textcolor{blue}{(-0.05)} \\ 
Group Convolution &79.95 \textcolor{red}{(-4.64)} &89.88 \textcolor{red}{(-2.34)} &88.86 \textcolor{red}{(-2.79)} &95.08 \textcolor{red}{(-0.66)} &14.56 \textcolor{blue}{(-32.95)} &29.05 \textcolor{blue}{(-9.11)} \\
Depthwise Convolution &73.64 \textcolor{red}{(-10.95)} &86.68 \textcolor{red}{(-5.54)} &84.82 \textcolor{red}{(-6.83)} &94.36 \textcolor{red}{(-1.38)} &13.24 \textcolor{blue}{(-34.27)} &25.22 \textcolor{blue}{(-12.94)} \\ 
\bottomrule
\end{tabular}
\end{table*}

\subsubsection{Analysis of Convolution Type in Dual Convolution Layers}In this part
of our ablation study, we focus on evaluating the impact of different convolution types used in the Dual Dynamic Convolution Layer
(DDC) Layer on the performance of ABCDWaveNet. We compare four convolution types: Standard Convolution, Dilated Convolution with a dilation rate of 2, Group Convolution with a group size of 4, and Depthwise Convolution. The results of this study are summarized in Table \ref{tab:conv_ablation_study}.
Dynamic Convolution demonstrates the most remarkable performance, attaining an IoU of 84.59 and an F1 Score of 92.22. Significantly, in the case where the GFLOPs of Dynamic Convolution are nearly equivalent to those of Standard Convolution and Dilated Convolution, it is respectively 2.26 and 5.10 higher than them. Although dilated convolution broadens the receptive field, it reduces the capability to capture fine spatial details, consequently degrading performance in scenarios that require complex feature extraction.

Group Convolution achieves comparatively good computational efficiency. However, this computational advantage is offset by performance degradation. Specifically, the IoU and F1 Score decrease by 4.64 and 2.34 points, respectively. The reduced parameterization impedes feature integration across channels, which makes it difficult for the model to capture complex spatial relationships that are crucial for high-precision tasks.
Depthwise Convolution presents the lowest accuracy, having an IoU of 73.64 and an F1 Score of 86.68. Although it attains the smallest numbers of parameters and GFLOPs, its accuracy level is insufficient for high-precision applications. Consequently, this limits its utility in complex scenarios that demand detailed feature extraction.
Overall, Dynamic Convolution stands out as the optimal choice, achieving high performance levels without a substantial increase in GFLOPs. It reliably maintains accuracy even in complex scenarios, such as foggy conditions where precise feature extraction is essential. In comparison, other convolution types face difficulties in sustaining these capabilities, underscoring the distinct advantages of Dynamic Convolution in challenging environments.

\subsection{Comparison With SOTA Methods}
\begin{figure}[htbp]
    \centering
    \includegraphics[width=0.5\textwidth]{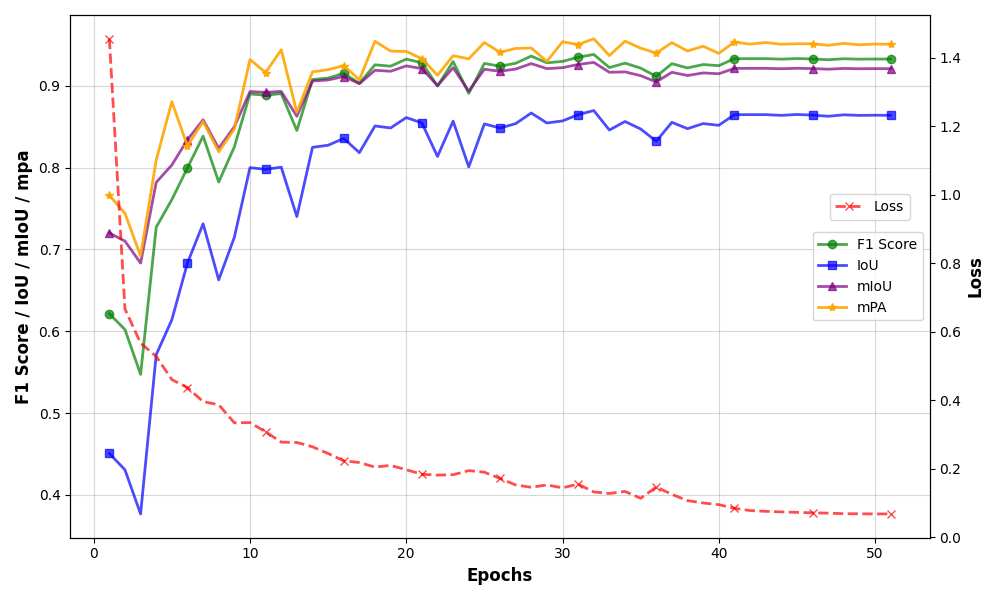}  
    \caption{The loss and performance metrics over training epochs on Foggy-Puddle dataset.}
    \label{fig:loss} 
\end{figure}

To evaluate the performance of ABCDWaveNet, we conducted a comprehensive comparison with eleven state-of-the-art methods characterized by their classical architectures, recent publications, and cutting-edge advancements in road ponding detection and related fields. The methods included F3Net \cite{f3net}, CCNet \cite{ccnet}, U2Net \cite{u2net}, SegFormer-B2 \cite{xie2021segformer}, SeaFormer-Base \cite{seaformer}, UDTransNet \cite{udtransnet}, and VWFormer \cite{vwformer} for general semantic segmentation; BWG \cite{bwg} for foggy semantic segmentation; and SWNet \cite{swnet}, HomoFusion \cite{homofusion}, and AGSENet \cite{agsenet} specifically for road ponding detection. To ensure fairness, we utilized publicly available code and conducted all experiments within a consistent environment, assessing all prediction maps using a unified codebase. The training process is shown in the Fig.\ref{fig:loss}

\begin{table*}[htbp]
\centering
\begin{threeparttable}
\caption{Comprehensive quantitative comparison of state-of-the-art (SOTA) methods on the Foggy-Puddle dataset. Metrics highlighted in \textcolor{blue}{\textbf{Bold Blue}} denote the best performance, while those in \textcolor{green}{\textbf{Bold Green}} indicate the second-best performance.}  
\label{foggy_methods}
\tabcolsep=0.25cm
\renewcommand\arraystretch{1.25}
\begin{tabular}{c|c|cccc|c|c|c}
\hline
\multirow{2}{*}{\textbf{Methods}} & \multirow{2}{*}{\textbf{Pub. Year}} & \multicolumn{4}{c|}{\textbf{Foggy-Puddle}} & \multirow{2}{*}{\textbf{\#Param (M)}} & \multirow{2}{*}{\textbf{GFLOPs}} & \multirow{2}{*}{\textbf{FPS}} \\
\cline{3-6}
 & & \textbf{IoU} & \textbf{F1 Score} & \textbf{MIoU} & \textbf{MPA} & & \\
\hline
F3Net\cite{f3net} & AAAI'2020 & 77.42 & 88.65 & 88.64 & 94.57 & 25.54 &9.48 & 31.97 \\
CCNet\cite{ccnet} & IEEE TPAMI'2020 & 79.55 & 89.15 & 89.78 & 95.20 & 71.27 & 79.69 &22.33 \\
U2Net\cite{u2net} & PR'2020 & 80.20 & 89.18 & 90.07 & 94.96 & 1.77 & 13.07 &25.51 \\
SegFormer-B2\cite{xie2021segformer} & NeurIPS'2021 & 78.26 & 88.47 & 89.17 & 94.86 & 27.45 & 14.98 &24.71\\
SWNet\cite{swnet} & IEEE TITS'2022 & 79.44 & 89.02 & 89.88 & 95.14 & 21.81 & 26.84&38.83\\
HomoFusion\cite{homofusion} & ICCV'2023 & 78.35 & 88.65 & 89.73 & 94.83 & 1.24 & 61.20 &25.41\\
SeaFormer-Base\cite{seaformer} & ICLR'2023 & 79.96 & 89.61 & 90.28 & 95.22 & 8.58 & 0.43 &42.68 \\
UDTransNet\cite{udtransnet} & Neural Netw'2024 & 76.56 & 88.11 & 86.72 & 94.59 & 33.93 & 34.58 & 18.64 \\
VWFormer\cite{vwformer} & ICLR'2024  & 76.73 & 88.16 & 86.83 & 92.78& 94.74 & 37.30 &10.60 \\
AGSENet\cite{agsenet} & TITS'2024 & 79.29 & 89.55 & 88.45 & 92.32 & 2.05 & 16.01 &22.74\\
BWG\cite{bwg} & AAAI'2024 & \textcolor{green}{\textbf{81.08}} & \textcolor{green}{\textbf{89.24}} & \textcolor{green}{\textbf{90.35}} & \textcolor{green}{\textbf{95.15}} & 73.46 & 93.54 & 15.52\\
\rowcolor{blue!8} \textcolor{blue}{\textbf{ABCDWaveNet (Ours)}} & \textcolor{blue}{\textbf{- - - -}} & \textcolor{blue}{\textbf{84.59}} & \textcolor{blue}{\textbf{92.22}} & \textcolor{blue}{\textbf{91.65}} & \textcolor{blue}{\textbf{95.74}} & 47.51 & 38.16 & 25.48 \\
\hline
\end{tabular}
\begin{tablenotes}
\small
\item Notes: We evaluate across multiple metrics: Intersection over Union (IoU), F1 Score, Mean Intersection over Union (MIoU), Mean Pixel Accuracy (MPA), model parameter count (\#Param in millions), computational complexity (GFLOPs), and inference speed (FPS) with an input image resolution of 256$\times$256. The FPS values were measured on the NVIDIA Jetson AGX Orin.
\end{tablenotes}
\end{threeparttable}
\end{table*}

\begin{table*}[htbp]
\centering
\begin{threeparttable}
\caption{Comprehensive quantitative comparison of state-of-the-art (SOTA) methods on the Puddle-1000 dataset. Metrics highlighted in \textcolor{blue}{\textbf{Bold Blue}} denote the best performance, while those in \textcolor{green}{\textbf{Bold Green}} indicate the second-best performance.}%evaluated across multiple metrics: Intersection over Union (IoU), F1 Score, Mean Intersection over Union (MIoU), Mean Pixel Accuracy (MPA), model parameter count (\#Param in millions), computational complexity (GFLOPs), and inference speed (FPS) with an input image resolution of 256$\times$256. The FPS values were measured on the NVIDIA Jetson AGX Orin. }
\label{PUDDLE-1000_methods}
\tabcolsep=0.25cm
\renewcommand\arraystretch{1.25}
\begin{tabular}{c|c|cccc|c|c|c}
\hline
\multirow{2}{*}{\textbf{Methods}} & \multirow{2}{*}{\textbf{Pub. Year}} & \multicolumn{4}{c|}{\textbf{Puddle-1000}} & \multirow{2}{*}{\textbf{\#Param (M)}} & \multirow{2}{*}{\textbf{GFLOPs}} & \multirow{2}{*}{\textbf{FPS}}\\
\cline{3-6}
 & & \textbf{IoU} & \textbf{F1 Score} & \textbf{MIoU} & \textbf{MPA} & & \\
\hline
F3Net\cite{f3net} & AAAI'2020 & 81.84 &  90.01 & 90.79  & 94.36  & 25.54 & 9.48 &31.97 \\
CCNet\cite{ccnet} & IEEE TPAMI'2020 & 82.23 & 91.03 & 90.73 & 94.40 & 71.27 & 79.69 &22.33\\
U2Net\cite{u2net} & PR'2020 & 83.57 & 91.22 & 90.68 &95.17  & 1.77 & 13.07 &25.51 \\
SegFormer-B2\cite{xie2021segformer} & NeurIPS'2021 & 80.49  & 89.19 & 90.12 & 94.55 & 27.45 & 14.98 &24.71\\
SWNet\cite{swnet} & IEEE TITS'2022 & 82.01 & 90.11 & 90.89 & 94.77 & 21.81 & 26.84 &38.83\\
HomoFusion\cite{homofusion} & ICCV'2023 & 80.26 & 89.13 &89.91 & 95.67 & 1.24 & 61.20 &25.41 \\
SeaFormer-Base\cite{seaformer} & ICLR'2023 & 81.22 & 89.19 & 90.35 & 94.98& 8.58 & 0.43 &42.68\\
UDTransNet\cite{udtransnet} & Neural Netw'2024 & 77.59 & 88.64 & 87.38 & 95.56 & 33.93 & 34.58 &18.64\\
VWFormer\cite{vwformer} & ICLR'2024 & 77.98 & 89.27 & 88.70 & 94.64 & 94.74 & 37.30 &10.60\\
BWG\cite{bwg} & AAAI'2024 & 83.94 & 91.27 & 91.35 & 95.23 & 73.46 & 93.54 &15.52\\
AGSENet\cite{agsenet} & TITS'2024 & \textcolor{green}{\textbf{84.09}} & \textcolor{green}{\textbf{91.36}} & \textcolor{green}{\textbf{91.96}} & \textcolor{green}{\textbf{95.32}} & 2.05 & 16.01 &22.74\\
\rowcolor{blue!8} \textcolor{blue}{\textbf{ABCDWaveNet (Ours)}} & \textcolor{blue}{\textbf{- - - -}} & \textcolor{blue}{\textbf{85.84}} & \textcolor{blue}{\textbf{93.06}} & \textcolor{blue}{\textbf{92.45}} & \textcolor{blue}{\textbf{96.89}} & 47.51 & 38.16 &25.48 \\
\hline
\end{tabular}
\begin{tablenotes}
\small
\item Notes: We evaluate across multiple metrics: Intersection over Union (IoU), F1 Score, Mean Intersection over Union (MIoU), Mean Pixel Accuracy (MPA), model parameter count (\#Param in millions), computational complexity (GFLOPs), and inference speed (FPS) with an input image resolution of 256$\times$256. The FPS values were measured on the NVIDIA Jetson AGX Orin.
\end{tablenotes}
\end{threeparttable}
\end{table*}

\begin{table*}[htbp]
\centering
\begin{threeparttable}
\caption{Comprehensive quantitative comparison of state-of-the-art (SOTA) methods on the Foggy low-light Puddle dataset. Metrics highlighted in \textcolor{blue}{\textbf{Bold Blue}} denote the best performance, while those in \textcolor{green}{\textbf{Bold Green}} indicate the second-best performance.}%, evaluated across multiple metrics: Intersection over Union (IoU), F1 Score, Mean Intersection over Union (MIoU), Mean Pixel Accuracy (MPA), model parameter count (\#Param in millions), computational complexity (GFLOPs), and inference speed (FPS) with an input image resolution of 256$\times$256. The FPS values were measured on the NVIDIA Jetson AGX Orin. }
\label{foggylowlight_methods}
\tabcolsep=0.25cm
\renewcommand\arraystretch{1.25}
\begin{tabular}{c|c|cccc|c|c|c}
\hline
\multirow{2}{*}{\textbf{Methods}} & \multirow{2}{*}{\textbf{Pub. Year}} & \multicolumn{4}{c|}{\textbf{Foggy Low-light Puddle}} & \multirow{2}{*}{\textbf{\#Param (M)}} & \multirow{2}{*}{\textbf{GFLOPs}}& \multirow{2}{*}{\textbf{FPS}} \\
\cline{3-6}
 & & \textbf{IoU} & \textbf{F1 Score} & \textbf{MIoU} & \textbf{MPA} & & \\
\hline
F3Net\cite{f3net} & AAAI'2020 & 55.44 & 72.49 & 72.68 & 81.01 & 25.54 & 9.48 &31.97 \\
CCNet\cite{ccnet} & IEEE TPAMI'2020 & 57.74 & 73.18 & 73.24 & 83.28 & 71.27 & 79.69 &22.33 \\
U2Net\cite{u2net} & PR'2020 & 59.62 & 74.70 & 77.15 & 83.58 & 1.77 & 13.07 &25.51 \\
SegFormer-B2\cite{xie2021segformer} & NeurIPS'2021 & 55.10 &71.05  & 74.16& 83.77 & 27.45 & 14.98 &24.71\\
SWNet\cite{swnet} & IEEE TITS'2022 & 51.29 & 67.80 & 71.79 & 82.55 & 21.81 & 26.84 &38.83\\
HomoFusion\cite{homofusion} & ICCV'2023 & 58.63 & 75.45 & 74.02 &83.98  & 1.24 & 61.20 &25.41\\
SeaFormer-Base\cite{seaformer} & ICLR'2023 &60.50  & 75.39 & 77.40 & 85.67 & 8.58 & 0.43 &42.68\\
UDTransNet\cite{udtransnet} & Neural Netw'2024 & 60.09 & 77.14 & 75.07 & 87.56 & 33.93 & 34.58 &18.64\\
VWFormer\cite{vwformer} & ICLR'2024 & 60.96 & 77.57 & 76.16 & 87.98 & 94.74 & 37.30 &10.60\\
AGSENet\cite{agsenet} & TITS'2024 & 59.11 & 74.60 & 76.81 & 83.19 & 2.05 & 16.01 &22.74\\
BWG\cite{bwg} & AAAI'2024 & \textcolor{green}{\textbf{61.48}} & \textcolor{green}{\textbf{77.83}} & \textcolor{green}{\textbf{76.32}} & \textcolor{green}{\textbf{88.23}} & 73.45 & 93.54 &15.52\\
\rowcolor{blue!8} \textcolor{blue}{\textbf{ABCDWaveNet (Ours)}} & \textcolor{blue}{\textbf{- - - -}} & \textcolor{blue}{\textbf{62.51}} & \textcolor{blue}{\textbf{78.84}} & \textcolor{blue}{\textbf{76.93}} & \textcolor{blue}{\textbf{88.66}} & 47.51 & 38.16 &25.48\\ 
\hline
\end{tabular}
\begin{tablenotes}
\small
\item Notes: We evaluate across multiple metrics: Intersection over Union (IoU), F1 Score, Mean Intersection over Union (MIoU), Mean Pixel Accuracy (MPA), model parameter count (\#Param in millions), computational complexity (GFLOPs), and inference speed (FPS) with an input image resolution of 256$\times$256. The FPS values were measured on the NVIDIA Jetson AGX Orin.
\end{tablenotes}
\end{threeparttable}
\end{table*}

\begin{figure*}[htbp]
    \centering
    \includegraphics[width=1\textwidth]{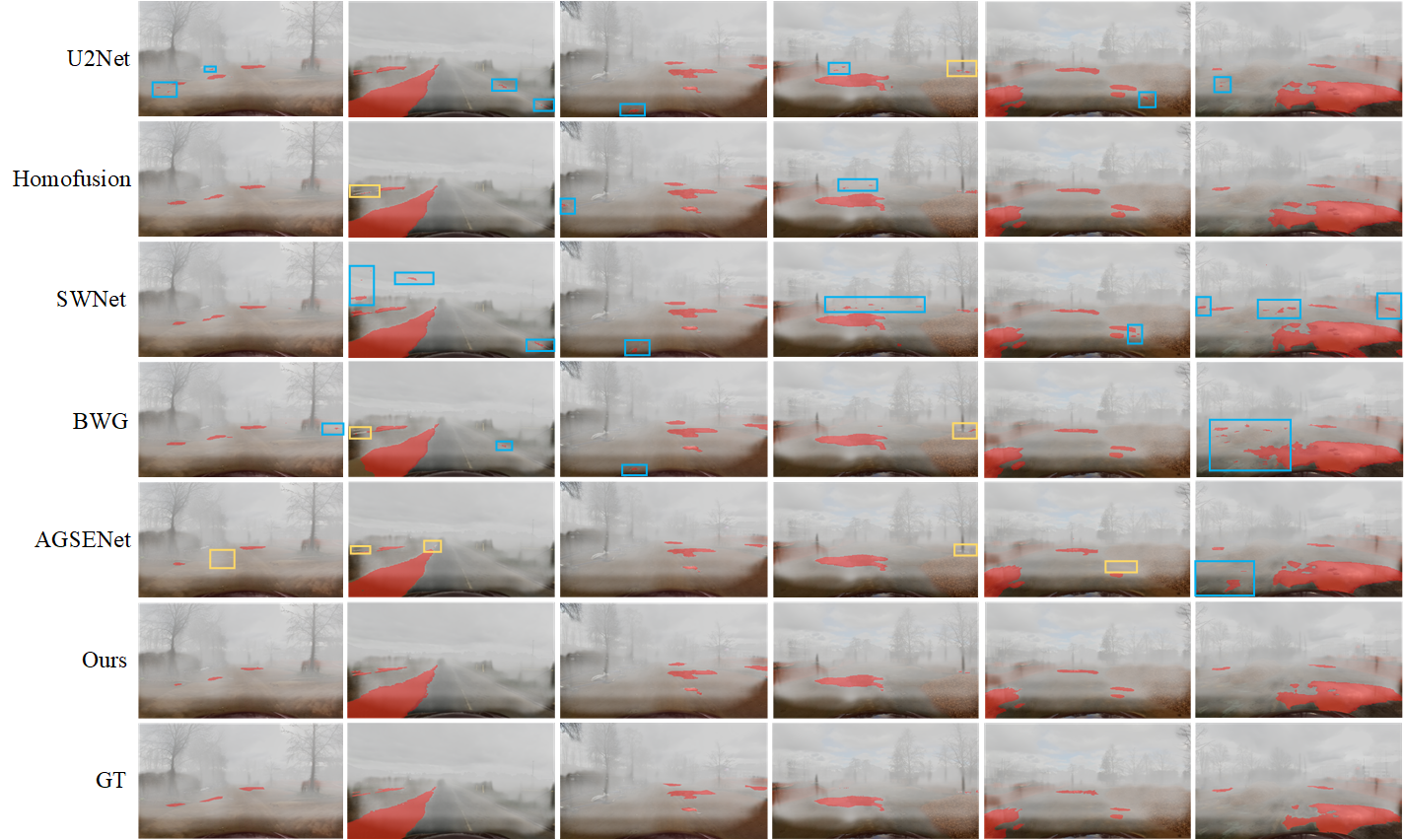}  
    \caption{Representative prediction results on the Foggy-Puddle dataset, comparing ABCDWaveNet (Ours), AGSENet, BWG, SWNet, Homofusion, and U2Net. The bottom row shows the ground-truth annotations. Yellow boxes indicate missed detections, while blue boxes indicate false detections.}
    \label{fig:keshihua} 
\end{figure*}

\subsubsection{Foggy-Puddle}
Table \ref{foggy_methods} provides a comprehensive evaluation of various methods on the Foggy-Puddle dataset. Our proposed ABCDWaveNet achieves state-of-the-art performance, significantly surpassing existing methods. Specifically, ABCDWaveNet attains an IoU of 84.59\%, outperforming the second-best method, BWG \cite{bwg}, by 3.51 percentage points, and an F1 Score of 92.22\%, exceeding BWG by 2.98 percentage points. Fig. \ref{fig:keshihua} further highlights the qualitative advantages of ABCDWaveNet. It demonstrates more precise segmentation with fewer missed detections (yellow boxes) and fewer false detections (blue boxes) compared to AGSENet, BWG, SWNet, Homofusion, and U2Net.

In addition to its superior accuracy, ABCDWaveNet delivers competitive efficiency, achieving an inference speed of 25.48 FPS on the NVIDIA Jetson AGX Orin. This combination of high accuracy and computational efficiency establishes ABCDWaveNet as a practical solution for ADAS early-warning systems in challenging weather conditions.

\subsubsection{Puddle-1000}
Table \ref{PUDDLE-1000_methods} presents the results on the Puddle-1000 dataset, representing road ponding segmentation under normal weather conditions. ABCDWaveNet achieves the highest IoU of 85.84\%, exceeding AGSENet \cite{agsenet}, the second-best method, by 1.75 percentage points. The F1 Score of ABCDWaveNet is 93.06\%, which is 1.70 percentage points higher than AGSENet. Additionally, ABCDWaveNet records an MIoU of 92.45\% and an MPA of 96.89\%, indicating improvements of 0.49 and 1.57 percentage points, respectively.

\subsubsection{Foggy Low-light Puddle}
To assess the robustness of ABCDWaveNet under more challenging conditions, we evaluated it on the Foggy Low-light Puddle dataset (Table \ref{foggylowlight_methods}). This dataset combines foggy and low-light conditions, presenting significant challenges for segmentation models. ABCDWaveNet achieves an IoU of 62.51\%, outperforming BWG \cite{bwg} by 1.03 percentage points. The F1 Score is 78.84\%, 1.01 percentage points higher than BWG. The MIoU and MPA are 76.93\% and 88.66\%, respectively, indicating consistent improvements over existing methods. 

The experimental results across all datasets indicate that ABCDWaveNet consistently outperforms existing state-of-the-art methods in road ponding segmentation tasks, especially under adverse weather conditions. The significant improvements in IoU and F1 Score suggest that our model effectively captures and differentiates the features of road ponding, even when visibility is compromised. While ABCDWaveNet has a moderate parameter count and computational cost, the gains in segmentation accuracy justify these resources, particularly for safety-critical applications like advanced driving assistance systems.

\begin{figure}[htbp]
    \centering
    \includegraphics[width=0.5\textwidth]{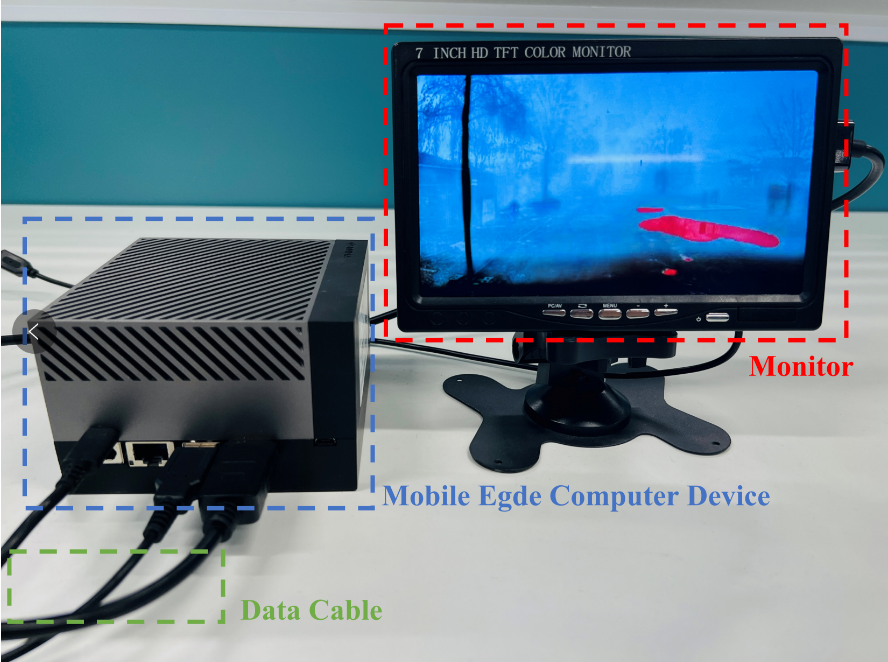}  
    \caption{A edge computing system built with NVIDIA Jetson AGX Orin.} %After experimental validation, it has been demonstrated that our method can run accurately on the vehicle’s edge computing platform with limited computational resources, achieving a performance of 25.48 FPS
    \label{fig:bushu} 
\end{figure}

\subsection{Model Deployment}
To extend the evaluation of our algorithm to real-world scenarios, we implemented and tested the model within an automotive environment in addition to utilizing existing datasets. The vehicle-mounted edge computing system, depicted in Fig. \ref{fig:bushu}, comprises components such as the NVIDIA Jetson AGX Orin, a display monitor and a data cable. As presented in Table \ref{foggy_methods}, our model achieved frame rates of 25.48 FPS. These evaluations confirm that our approach reliably delivers high accuracy and operational efficiency on the vehicle’s edge platform, thereby meeting the demands of ADAS early-warning systems despite resource limitations.

\section{Conclusion}
In this work, we proposed ABCDWaveNet, a novel framework for robust road ponding detection under foggy conditions, leveraging a synergistic integration of dynamic convolution and wavelet-based frequency-spatial enhancement. Through the Bidomain Information Synergy module and Aggregation-Broadcast-Coupling mechanism, the model effectively captures both global structure and fine details across scales. Extensive experiments on Puddle-1000, Foggy-Puddle and newly proposed Foggy Low-Light Puddle datasets demonstrate state-of-the-art performance on edge devices, validating the method’s practicality for ADAS applications. We hope that this work will offer practical insights for advancing robust and proactive road safety systems under degraded visual conditions. 

\textbf{\textcolor{blue}{To support future research and promote reproducibility, we are pleased to openly share our code and datasets with the community.}}

%This study introduced ABCDWaveNet, a novel deep learning framework specifically designed for road ponding detection under adverse foggy weather conditions. By seamlessly integrating dynamic convolution, wavelet-based feature extraction, and the Aggregation-Broadcast-Coupling mechanism, ABCDWaveNet achieves robust multi-scale spatial-frequency representation. The proposed architecture demonstrated superior performance on challenging datasets, achieving significant improvements in Intersection over Union (IoU) and F1 Score compared to existing methods.

%The experimental results validated the efficacy of ABCDWaveNet, highlighting its robustness and accuracy in low-visibility environments. Future research could explore the adaptation of ABCDWaveNet to other adverse weather scenarios or extend its application to additional environmental perception tasks. By addressing the critical challenge of accurate road ponding detection in difficult conditions, this study contributes to the advancement of Advanced  Driver Assistance Systems and supports safer driving experiences across diverse weather environments.

%\section*{Authors’ contribution}
%All authors have jointly undertaken and contributed to the following tasks during the research process, which include but are not limited to: (1) For the prototype and theoretical developments including system or experimental design of this article. (2) Contribution to drafting the article. (3) Approval of the final submit version of the article. 
\section*{Compliance with ethics guidelines}
Ronghui Zhang, Dakang Lyu, Tengfei Li, Yunfan Wu, Ujjal MANANDHAR, Benfei Wang, Junzhou Chen, Bolin
Gao, Danwei Wang, and Yiqiu Tan declare that they have no conflict of interest or financial conflicts to disclose.
%\printcredits

%% Loading bibliography style file
%\bibliographystyle{model1-num-names}
%\bibliographystyle{cas-model2-names}
%\bibliographystyle{unsrt}
\bibliographystyle{elsarticle-num-names}

% Loading bibliography database
\bibliography{reference}

%\bio{figs/cas-pic1}
%Author biography with author photo.biography.
%\endbio
\end{document}